\documentclass[journal]{IEEEtran}
\def\mathbi#1{\textbf{\em #1}}

\usepackage{cite}

\ifCLASSINFOpdf\usepackage{multirow}
\usepackage{amsmath} \usepackage{booktabs} \usepackage{amsfonts}
\usepackage{amsmath} \usepackage{amsthm} \usepackage{soul} \soulregister\cite7
\soulregister\citep7 \soulregister\citet7 \soulregister\ref7
\soulregister\pageref7

\usepackage{color}

\usepackage[caption=false]{subfig}
\usepackage[ruled,linesnumbered]{algorithm2e} \usepackage[pdftex]{graphicx}
\usepackage{url} 

\begin{document}
\title{Fine-Grained Trajectory-based Travel Time Estimation for Multi-city Scenarios Based on \\Deep Meta-Learning}

\author{Chenxing~Wang,~\IEEEmembership{Member,~IEEE,} Fang~Zhao, Haichao Zhang, Haiyong~Luo, ~\IEEEmembership{Member,~IEEE,} \\ Yanjun
Qin, and Yuchen Fang
\thanks{This manuscript has been accepted in IEEE Transactions on Intelligent Transportation Systems for future publication.

\textsuperscript{\textcopyright} 2022 IEEE.  Personal use of this material is permitted.  Permission from IEEE must be obtained for all other uses, in any current or future media, including reprinting/republishing this material for advertising or promotional purposes, creating new collective works, for resale or redistribution to servers or lists, or reuse of any copyrighted component of this work in other works.}
}


\markboth{IEEE TRANSACTIONS ON INTELLIGENT TRANSPORTATION
 SYSTEMS} %
 {Wang \MakeLowercase{\textit{et al.}}: Fine-Grained Trajectory-based Travel Time Estimation for Multi-City Scenarios Based on Deep Meta-Learning}



\maketitle

\begin{abstract} \underline{T}ravel \underline{T}ime 
  \underline{E}stimation (TTE) is indispensable in intelligent transportation
system (ITS). It is significant to achieve the fine-grained
\underline{T}rajectory-based \underline{T}ravel \underline{T}ime
\underline{E}stimation (TTTE) for multi-city scenarios, namely to accurately
estimate travel time of the given trajectory for multiple city scenarios.
However, it faces great challenges due to complex factors including dynamic
temporal dependencies and fine-grained spatial dependencies. To tackle these
challenges, we propose a meta learning based framework, MetaTTE, to
continuously provide accurate travel time estimation over time by leveraging
well-designed deep neural network model called DED, which consists of
\underline{D}ata preprocessing module and
\underline{E}ncoder-\underline{D}ecoder network module. By introducing meta
learning techniques, the generalization ability of MetaTTE is enhanced using
small amount of examples, which opens up new opportunities to increase the
potential of achieving consistent performance on TTTE when traffic conditions
and road networks change over time in the future. The DED model adopts an
encoder-decoder network to capture fine-grained spatial and temporal
representations. Extensive experiments on two real-world datasets are
conducted to confirm that our MetaTTE outperforms six state-of-art baselines,
and improve 29.35\% and 25.93\% accuracy than the best baseline on Chengdu and
Porto datasets, respectively.
\end{abstract}

%


\section{Introduction}
\label{sec:one}
\IEEEPARstart{T}{ravel} time estimation (TTE) plays a vital role in mobile
navigation~\cite{amirian2016predictive}, route planning~\cite{bast2016route} and
ride-hailing services~\cite{wang2018learning,fu2020compacteta}. It is reported
that, Baidu Maps, which is one of the largest mobile map applications, has over
340 million monthly active users worldwide by the end of December 2016
~\cite{fang2020constgat}. In order to estimate travel time for users who desires
to know the traffic condition in advance and wisely plan their upcoming trips,
it is significant to develop a TTE model which is able to provide accurate
travel time estimation over time in real applications.

In this paper, we focus on the fine-grained end-to-end
\underline{T}rajectory-based \underline{T}ravel \underline{T}ime
\underline{E}stimation (TTTE) for multi-city scenarios. Given historical
trajectory data in multiple cities, our objective is to train single model and
provide accurate travel time estimation over time for all city scenarios. Please
note that it is different from other end-to-end trajectory-based methods, which
either heavily rely on road networks of specific cities
~\cite{zhang2018deeptravel,wang2018learning,fu2019deepist,li2019learning,fu2020compacteta,fang2020constgat,
zhang2020real} or require careful preprocessing for historical trajectories of
specific cities~\cite{wang2018will,qiu2019nei,lan2019deepi2t}.

\begin{figure}[h]
  \centering \includegraphics[width=7.5cm]{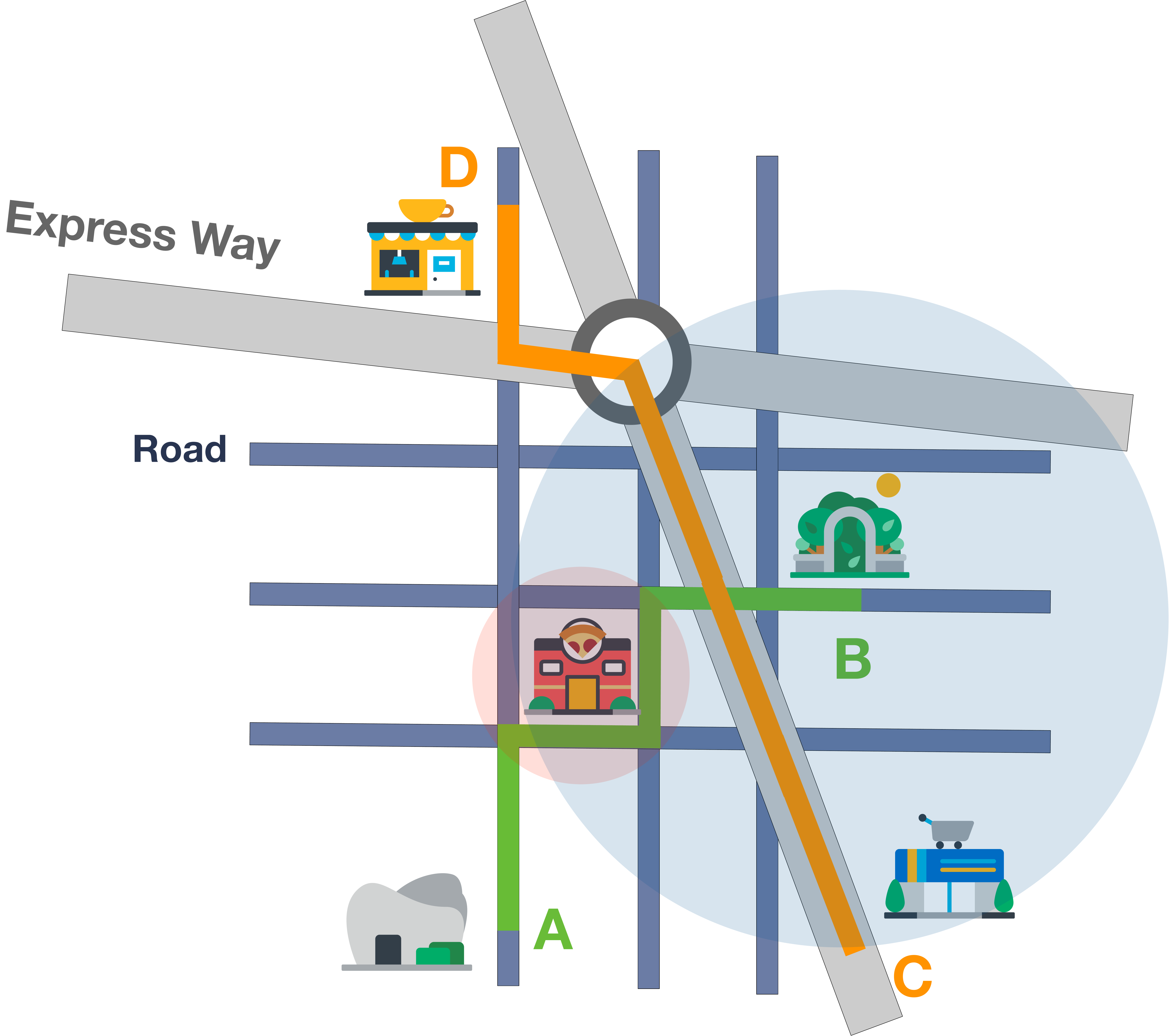}
  \caption{Two types of common scenarios in route planning applications.
  Route $A\rightarrow B$ contains relatively short distance gaps between
  adjacent key GPS points, while route $C\rightarrow D$ contains long distance
  gaps between adjacent key GPS points.}\label{fig:deeptte}
\end{figure}

However, it faces great challenges to provide accurate travel time estimation in
real applications over time, as the TTTE is affected by many complex factors:
\begin{itemize}

\item \textbf{Dynamic temporal dependencies.} Travel time estimation is
influenced by complex temporal factors, including the time varying traffic
conditions and evolving road networks. On one hand, traffic conditions which
implicitly influence travel time estimation change over time (i.e.~peak and
off-peak hours in a day, different days in a week etc.). On the other hand, road
networks that impacts the travel time are varying when roadworks are undertaken,
or streets are temporarily closed due to emergencies or regional restrictions
etc., which frequently happens in large cities. Some TTTE studies
~\cite{zhang2018deeptravel,wang2018learning,fu2019deepist,li2019learning,fu2020compacteta,fang2020constgat,
zhang2020real} utilize road network and traffic condition information in
different cities and provide travel time estimations with different models,
which performs well on their datasets. However, since the traffic conditions
and road networks dynamically change over time, travel time estimation model is
required to quickly fit latest traffic data to achieve satisfied performance
consistently. The aforementioned methods heavily rely on large amount of
historical traffic data which limits these models to continuously provide
accurate estimation in a fast learning scheme.

\item \textbf{Fine-grained spatial dependencies.} Some TTTE studies
~\cite{wang2018will,qiu2019nei,lan2019deepi2t} carefully preprocess trajectories
when providing travel time estimation using raw trajectory data. DeepTTE
~\cite{wang2018will}, for instance, resamples each trajectory data such that the
distance gap between two consecutive points are relatively within the same range
(i.e. 200 to 400 meters). However, the fine-grained spatial dependencies in
daily scenarios whose distance gap ranges are less than 200 meters or more than
400 meters are ignored. As illustrated in Figure~\ref{fig:deeptte}, we assume
that there are two users using route planning applications to navigate
themselves to the destination, with the route $A\rightarrow B$ and route
$C\rightarrow D$, respectively. For route $A\rightarrow B$, there are many
crossings in the whole trajectory which may contains several key GPS points (at
the crossing) with 50 to 100-meter distance gap between neighboring points. The area masked with smaller red circle contains short-term features (i.e.
the travel speed of the vehicles is slower and the traffic condition is
relatively bad) of the road segments with less than 200-meter distance gaps
which cannot be captured by DeepTTE For route $C\rightarrow D$, the whole
trajectory may contain some key GPS points (at the crossing) with long distance
gaps on express way which may exceeds 2 kilometers. The area masked with larger
grey circle contains long-distance features (i.e.~the travel speed of the
vehicles is faster and the traffic condition is relatively good) corresponding
to the road segments with more than 400-meter distance gaps, which cannot also
be captured by DeepTTE either.
\end{itemize}

Therefore, it's necessary to design a novel fine-grained model to simultaneously
capture the dynamic temporal dependencies and fine-grained spatial dependencies,
which is able to continuously provide accurate travel time estimation over time.
Most existing works which utilizing conventional deep learning techniques ignore
the fine-grained spatial dependencies and is limited to provide accurate travel
time estimation in specific city at current time period as mentioned before.

\textit{To the best of our knowledge, this is the first work that investigates
the potential of the fine-grained TTTE to continuously provide accurate travel
time estimations for multi-city scenarios over time in a fast learning manner
based on meta learning.} To address the above challenges, we mainly make the
following contributions:
\begin{itemize}

\item We construct two TTE-Tasks from two real-world datasets for training,
corresponding to Chengdu, China and Porto, Portugal and divide the whole dataset
into different sub-datasets which contains regional trajectories. Since every
trajectory data is restricted to one unique region, MetaTTE is capable of
quickly fitting latest trajectory data by simply adding a task using single
model instead of training multiple models for multi-city scenarios which limits
the ability of fast learning.

\item We propose a novel deep neural network model called DED in the meta
learning based framework (MetaTTE), which consists of two modules: (i)
\underline{D}ata preprocessing module to remove the biases of outliers in
real-world datasets and transform data into proper forms; (ii)
\underline{E}ncoder-\underline{D}ecoder network module to firstly embed the
spatial and temporal attributes into spatial, short-term and long-term
embeddings, capture fine-grained spatial dependencies using recurrent neural
network (RNN), fuse high level features using attention mechanism and finally
estimate the travel time for multi-city scenarios. \item Moreover, to tackle the
dynamic temporal dependencies, we introduce meta learning techniques into travel
time estimation, which opens up new opportunities to continuously provide
accurate travel time estimations over time, especially to increase the potential
of achieving consistent performance when traffic conditions and road networks
change over time in the future.

\item By introducing meta-learning techniques, we enhance the generalization
ability of MetaTTE using small amount of data from different TTE-Tasks. This
enables our model to \textit{learn to learn}, which increases the potential of
accurate estimation in the future and decreases the time consumption to achieve
the goal of fast learning.

\item We conduct extensive experiments on two real-world large scale datasets
collected in Chengdu, China and Porto, Portugal. The evaluation results show
that our MetaTTE outperforms other state-of-art baselines. The source codes of
MetaTTE are publicly available at
Github\footnote{\url{https://github.com/morningstarwang/MetaTTE}}.
\end{itemize}

The remaining of this paper is organized as follows: Section~\ref{sec:two}
introduces the preliminary. Section~\ref{sec:three} presents the data
description and analysis. Section~\ref{sec:four} investigates our proposed
MetaTTE and technical details of TTTE.\@ Section~\ref{sec:five} presents empirical
studies. Then related works are discussed in Section ~\ref{sec:six}. Finally,
Section~\ref{sec:seven} concludes the paper.

\section{Preliminary}\label{sec:two}

In this section, we firstly introduce meta learning to provide a comprehensive
motivations for utilizing meta-learning techniques in TTTE and then present
notations and the objective of our proposed MetaTTE.

 \begin{table*}[!t] %
\captionsetup{justification=centering, labelsep=newline} \caption{The overview
of three meta-learning categories, i.e., metric-based, model-based and
optimization-based techniques and their main pros and cons. In this figure,
$k_{\theta}(\mathbi{x}, \mathbi{x}_i)$ is a kernel function which returns the
similarity between the two inputs $\mathbi{x}$ and $\mathbi{x}_i$, $y_i$ are
ground-truths for known inputs $\mathbi{x}_i$, $\theta$ are base-learner
parameters, and $g_{\phi}$ is a learnt optimizer with parameters $\phi$.}
\label{tab:meta-learning} \centering \begin{tabular}{l|l|l|l} \toprule & Metric
& Model & Optimization\\ \midrule \textbf{Key idea} & Input similarity &
Internal task representation & Optimize for fast adaptation \\
$P_{\theta}(\mathcal{Y}|\mathbi{x}, \mathcal{D}_{\mathcal{T}_j}^{train})$ &
$\Sigma_{(\mathbi{x}_i,\mathbi{y}_i) \in \mathcal{D}_{\mathcal{T}_j^{train}}}
k_{\theta}(\mathbi{x}, \mathbi{x}_i)\mathbi{y}_i$ & $f_{\theta}(\mathbi{x},
\mathcal{D}_{\mathcal{T}_i}^{train})$ & $
f_{g_{\phi}(\theta,\mathcal{D}_{\mathcal{T}_j}^{train})}(\mathbi{x}) $\\
\textbf{Pros} & +Simple and effective & +Flexible & +Generalization ability\\
\textbf{Cons} & -Limited to supervised learning & -Weak generalization &
-Computationally expensive \\ \bottomrule \end{tabular} \end{table*} 

\begin{algorithm}
  \SetNlSty{textbf}{}{:}  \SetKwInOut{Input}{Input}  \SetKwInOut{Output}{Output}
   \Input{$p(\mathcal{T})$: distribution over tasks}  \Input{$\alpha$, $\beta$:
  step size of the optimizers}  \Output{A partition of the bitmap} \BlankLine
  randomly initialize $\theta$\; \While{not done}{ Sample task $\mathcal{T}_i$,
  corresponding to loss $\mathcal{L}_{\mathcal{T}_i}$ on parameters
  $\theta_{\mathcal{T}_i}$ \; Compute
  $\theta_{\mathcal{T}_i}=U_{\mathcal{T}_i}^k(\theta)$, denoting to $k$ steps of
  SGD or Adam \; Update $\theta \leftarrow \theta +
  \epsilon(\theta_{\mathcal{T}_i} - \theta)$ } \caption{Reptile for supervised
  regression problem.}\label{algo_reptile}
  \end{algorithm}
\textbf{Meta Learning.} Learning quickly is a hallmark of human intelligence, whether it involves recognizing
objects from a few examples or quickly learning new skills after just minutes of
experience~\cite{finn2017model}. Naturally, we want our artificial agents to
perform like this. However, it is challenging to complete many tasks utilizing
deep learning methods, which generally needs far more data to reach the same
level of performance as humans do. Under such circumstances, meta-learning has
been recently suggested as one strategy to overcome this challenge
~\cite{naik1992meta}. The key idea for meta-learning agents is to improve their
learning ability time to time, or equivalently, learn to learn. Existing works
can be categorized into: metric-based methods, model-based methods and
optimization-based methods~\cite{huisman2020survey}. The overview of these
methods is shown in Table~\ref{tab:meta-learning}. Metric-based methods
learn feature space that can be used to compute predictions based on input
similarity scores. The concept of this type of methods is simple and they can be
fast at test phase when tasks are small. However, when tasks become larger (e.g.
amount of trajectory data), the pair-wise comparisons may become computationally
expensive. Model-based methods may not perform well when presented with larger
datasets~\cite{hospedales2020meta} and generalize less well than
optimization-based methods~\cite{finn2017meta}. Hence, we choose the
optimization-based methods because most of them are model-agnostic meta-learning
algorithms and their generalization ability and the scalability are reasonable.
There are two state-of-art optimization-based methods called MAML
~\cite{finn2017model} and Reptile~\cite{nichol2018first}. The limitations of the
former for travel time estimation lie in two aspects. On one hand, it's too
cumbersome that relies on higher-order gradients. The inner gradient step has to
be implemented manually (e.g. TensorFlow 2.x),  which is inconvenient for
problems which requires a large number of gradient steps. On the other hand,
MAML requires a train-test split for each task which is common for few-shot
learning tasks, whereas the problem settings in our travel time estimation are
more likely to conventional regression or classification problems, which
requires either the training data or test data in each task. To tackle these
problems, we choose Reptile in this paper, which can better meet our
requirements. Reptile works by repeatedly sampling a task, training on it and
moving the initialization towards the trained parameters on that task and has
achieved fair results on some well-established benchmarks for classification.
The optimization algorithm of Reptile is shown in Algorithm~\ref{algo_reptile}.
In this algorithm, $U_{\mathcal{T}}^k(\theta)$ is the function which updates
parameters $\theta$ $k$ times using $k$ new batches of data sampled from task
$\mathcal{T}$ and $\epsilon$ is the learning rate for parameters' update. Notice
that in the last step,we treat $\theta - \theta_{\mathcal{T}_i}$ as the gradient
and adopt an adaptive algorithm for the update. Since Reptile can obtain fair
results on most of the benchmarks compared with MAML and it simplifies both the
implementation and the experimental settings which are key factors in real
applications, we utilize Reptile to be our base training algorithm to train
MetaTTE and the implementation details of our optimization algorithm is
presented in Section~\ref{optimize}.

\textbf{MetaTTE Trajectory.} We define a MetaTTE Trajectory $G$ from its
starting GPS coordinates $(p_1^1, p_2^1)$ to its destination $(p_1^n, p_2^n)$ as
a series of $n$ data points described by their GPS coordinate difference
attributes $(\Delta p_1,\Delta p_2)$, timestamp $t$, temporal attributes in
long-term $w$ (the day of week), and those in short-term $h$ (the hour of day).
Then MetaTTE trajectory $G$ is formulated as: \begin{equation} G =
\begin{bmatrix} \Delta p_1^1 & \Delta p_2^1 & w^1 & h^1 & t^1\\ \Delta p_1^2 &
\Delta p_2^2 & w^2 & h^2 & t^2\\ \cdots & \cdots & \cdots & \cdots & \cdots\\ \Delta p_1^{l} & \Delta
p_2^{l} &  w^l & h^l & t^l \end{bmatrix} \end{equation} where $l=n-1$
represents the number of data points in the MetaTTE trajectory $G$, $\Delta
p_1^j$ and $\Delta p_2^j$ $( j=1,2, \ldots, n-1)$ represent the difference of
latitudes and longitudes between $p_1^{j+1}$, $p_1^j$ and $p_2^{j+1}$, $p_2^j$,
respectively. We regard the first point $(\Delta p_1^1,\Delta p_2^1, w^1, h^1,
t^1)$ as the origin of this trajectory and the last point $(\Delta p_1^l,\Delta
p_2^l, w^l, h^l, t^l)$ as the destination. Hence we calculate the time
difference using $|t^l - t^1|$ as the label in our regression problem.

\textbf{\underline{T}ravel \underline{T}ime \underline{E}stimation
\underline{Task} (TTE-Task).} We define two tasks corresponding to Chengdu and
Porto, which are formulate as:
$\mathcal{T}_i=(\mathcal{D}_{\mathcal{T}_i}^{train},\mathcal{D}^{val},\mathcal{D}^{test})$
where $\mathcal{T}_i$ is the $i^{th}$ task,
$\mathcal{D}_{\mathcal{T}_i}^{train}, \mathcal{D}^{val}$ and $
\mathcal{D}^{test}$ are the train dataset for task $\mathcal{T}_i$, validate
dataset and test dataset for all TTE-Tasks, respectively.

\textbf{Objective.} Given $n$ TTE-Tasks, we train single deep learning model
using meta-learning techniques to learn the feature representations for each
task as well as the implicit dependencies between all TTE-Tasks. Therefore, the
objective is to estimate the total travel time for the query path belonging to
any task using the MetaTTE.\@

In real application scenarios, the route planning software or platform may
provide the key GPS locations along the query path, and MetaTTE is qualified to
estimate the total travel time continuously.

\section{Data Description and Preliminary Analysis}~\label{sec:three}

\subsection{Description}

Following two real-world datasets collected from two different regions in the
world are utilized to evaluate the performance of MetaTTE and the descriptions
and statistics of two datasets are shown in Table \ref{tab:stat_datasets}.

\begin{itemize}

\item{\textbf{Chengdu~\cite{dcjingsai}:} the Chengdu dataset is collected from
real-world taxis in Chengdu, China dated from Aug 3rd, 2014 to Aug 29th, 2014.
Over 1.4 billion GPS records are collected and over 14,000 taxis are involved.
We do not re-sample the GPS trajectory to a relatively fixed pattern (e.g.
60-second time gap between two GPS points~\cite{fu2019deepist}) but to keep the
original pattern (long time gap or short time gap) for each trajectory data in
the dataset.}

\item{\textbf{Porto~\cite{ecmlpkdd}:} the Porto dataset describes a complete
year (from Jul 1st, 2013 to Jun 30th, 2014) of the trajectories for all the 442
taxis running in Porto, Portugal. All the taxis are operated through a taxi
dispatch central using terminals installed in the vehicles. We remove all the
incomplete trajectories and calculate the total travel time for each
trajectory.}

\end{itemize}

\subsection{Analysis}

\subsubsection{Data Preprocessing}
\begin{table}
  \caption{The description and statistics of datasets.}~\label{tab:stat_datasets} \centering \begin{tabular}{l|l|l} \toprule Dataset &
  Chengdu & Porto\\ \midrule Travel Time Standard Deviation & 731.97sec &
  347.48sec \\ Travel Time Mean & 877.98sec & 691.29sec\\ The Number of
  Trajectories & 1,540,438 & 1,674,152\\ \bottomrule
\end{tabular}
\end{table}
We first sample all the available trajectories in these datasets and convert all
timestamps to seconds. We then divide each dataset into three parts, including
training data \textbf{(70\%)}, validation data \textbf{(10\%)} and test data
\textbf{(20\%)}. Notice that we  choose different date ranges of the historical
trajectory data for the training, validation and test part, respectively. More specifically, we
select trajectory data from August 3rd, 2014 to August 16th, 2014 to be the
training data for Chengdu dataset, from August 21st, 2014 to August 22nd, 2014 to be the validation data, and from August 24th, 2014 to August 29th, 2014 to be the test data for Chengdu dataset. Meanwhile, we select
trajectory data from July 1st, 2013 to February 28th, 2014 to be the training
data for Porto dataset, from March 1st, 2014 to April 1st, 2014 to be the validation data and from May 1st, 2014 to July 1st, 2014 to be the test data for Porto dataset. The mentioned procedures are
reproduced on all the baselines in this paper. 
\subsubsection{Data Analysis} 
\begin{figure} \centering \subfloat[Distribution of travel time of Chengdu
  dataset.]{\includegraphics[width=1.6in]{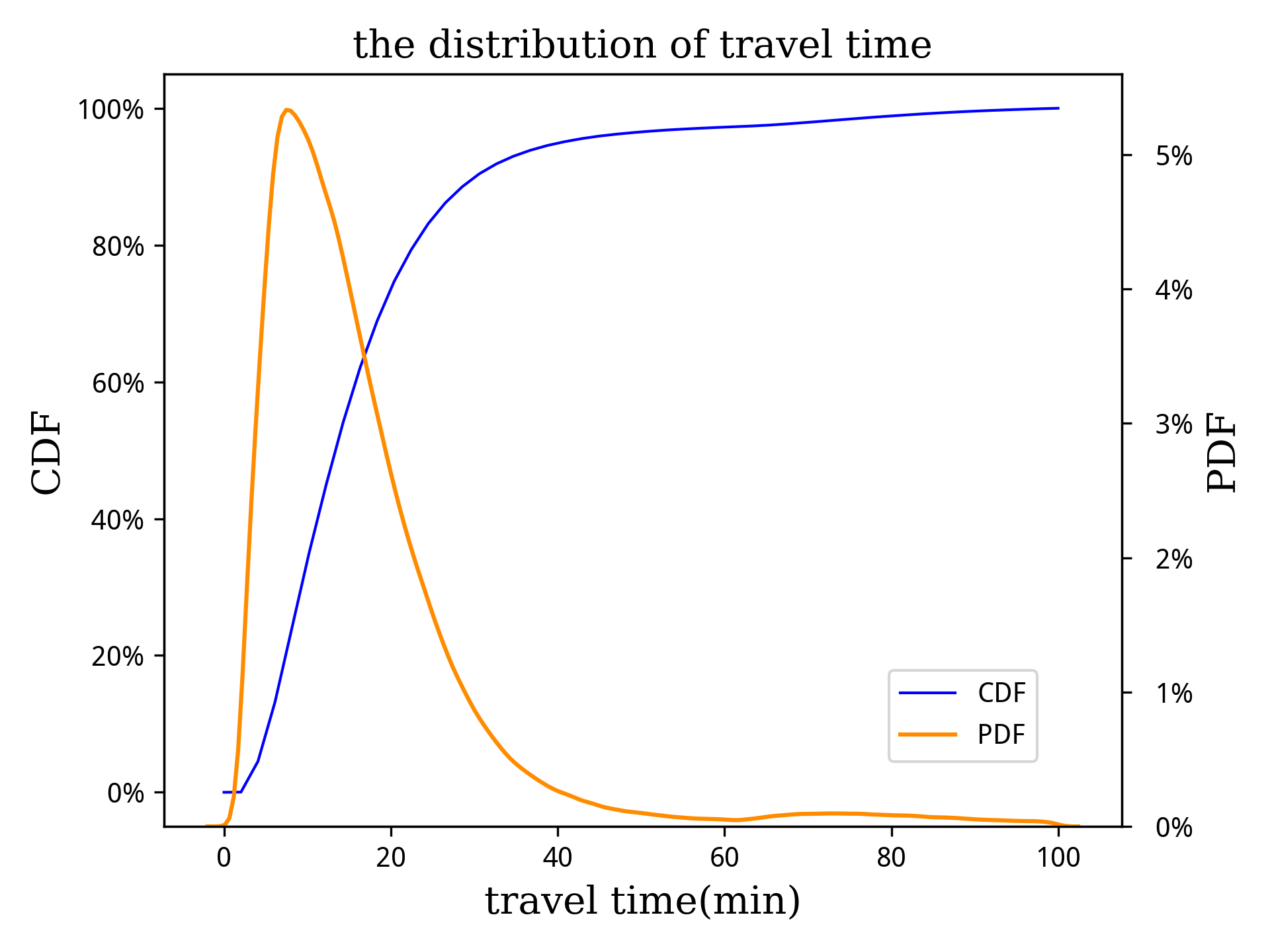}
  \label{fig:cdf_chengdu_travel_time_1} } \hfil \subfloat[Distribution of travel
  time of Porto
  dataset.]{\includegraphics[width=1.6in]{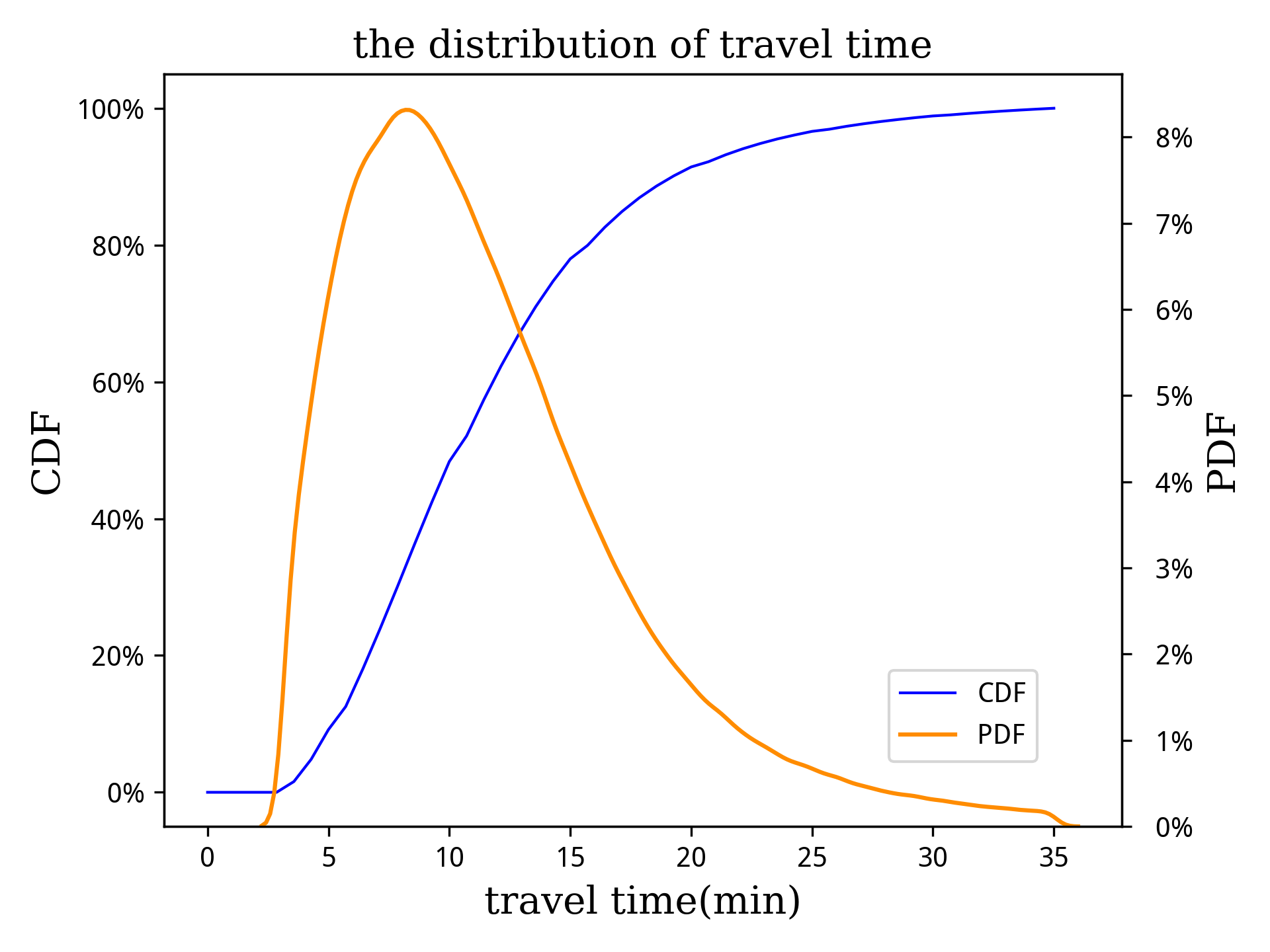}
  \label{fig:cdf_chengdu_travel_time_2} } \caption{Distribution of travel time.}~\label{fig:cdf_travel_time} \end{figure}
  
  \begin{figure} \centering \subfloat[Distribution of travel distance of
  Chengdu dataset.]{
  \includegraphics[width=1.6in]{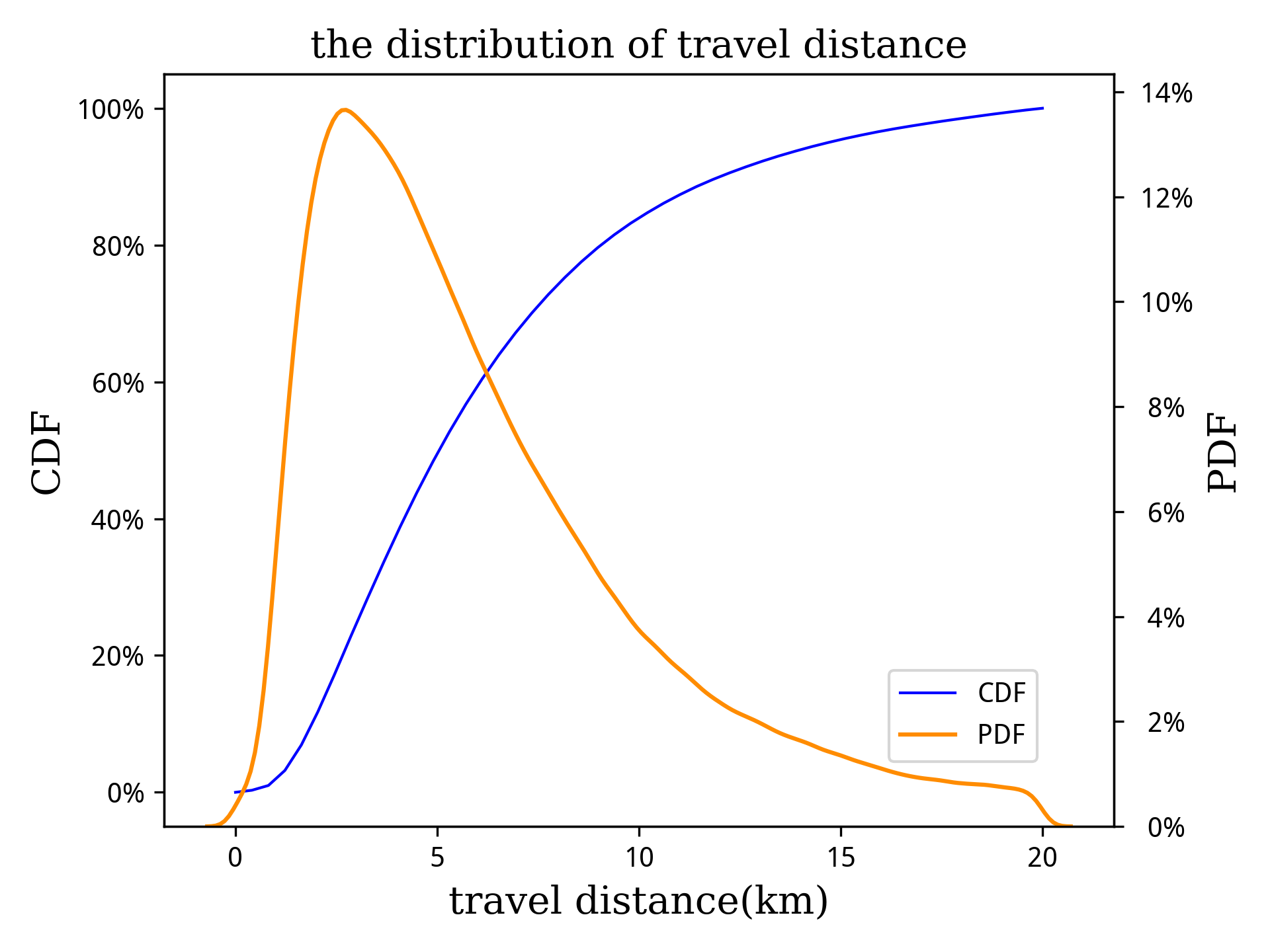}
  \label{fig:cdf_travel_distance_1} } \hfil \subfloat[Distribution of travel
  distance of Porto
  dataset.]{\includegraphics[width=1.6in]{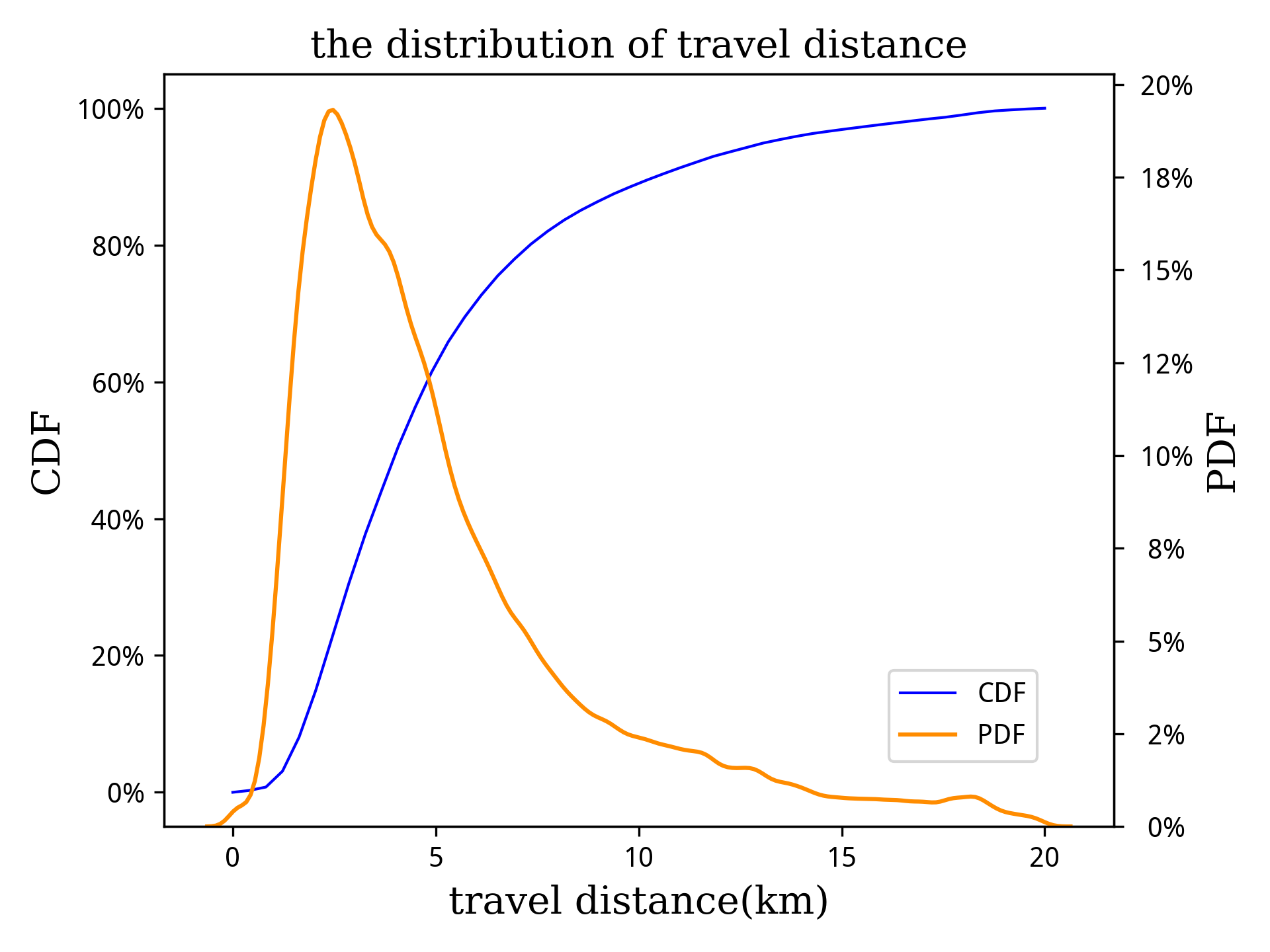}
  \label{fig:cdf_travel_distance_2} } \caption{Distribution of travel distance.}
~\label{fig:cdf_travel_distance} \end{figure}
  
  \begin{figure} \centering \subfloat[Chengdu
  dataset.]{\includegraphics[width=1.6in]{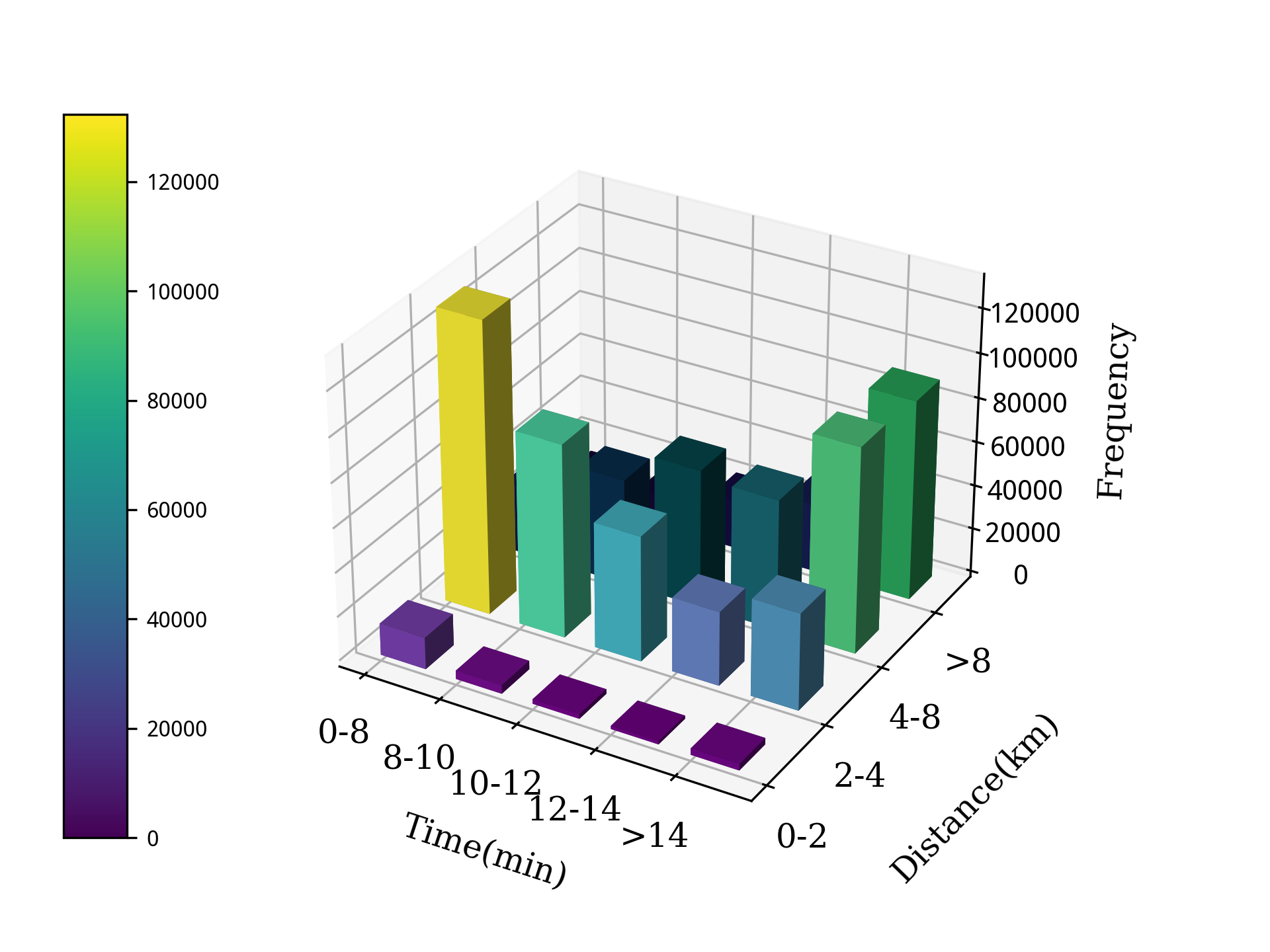}
  \label{fig:time_dist_chengdu_porto_1} } \hfil \subfloat[Porto
  dataset.]{\includegraphics[width=1.6in]{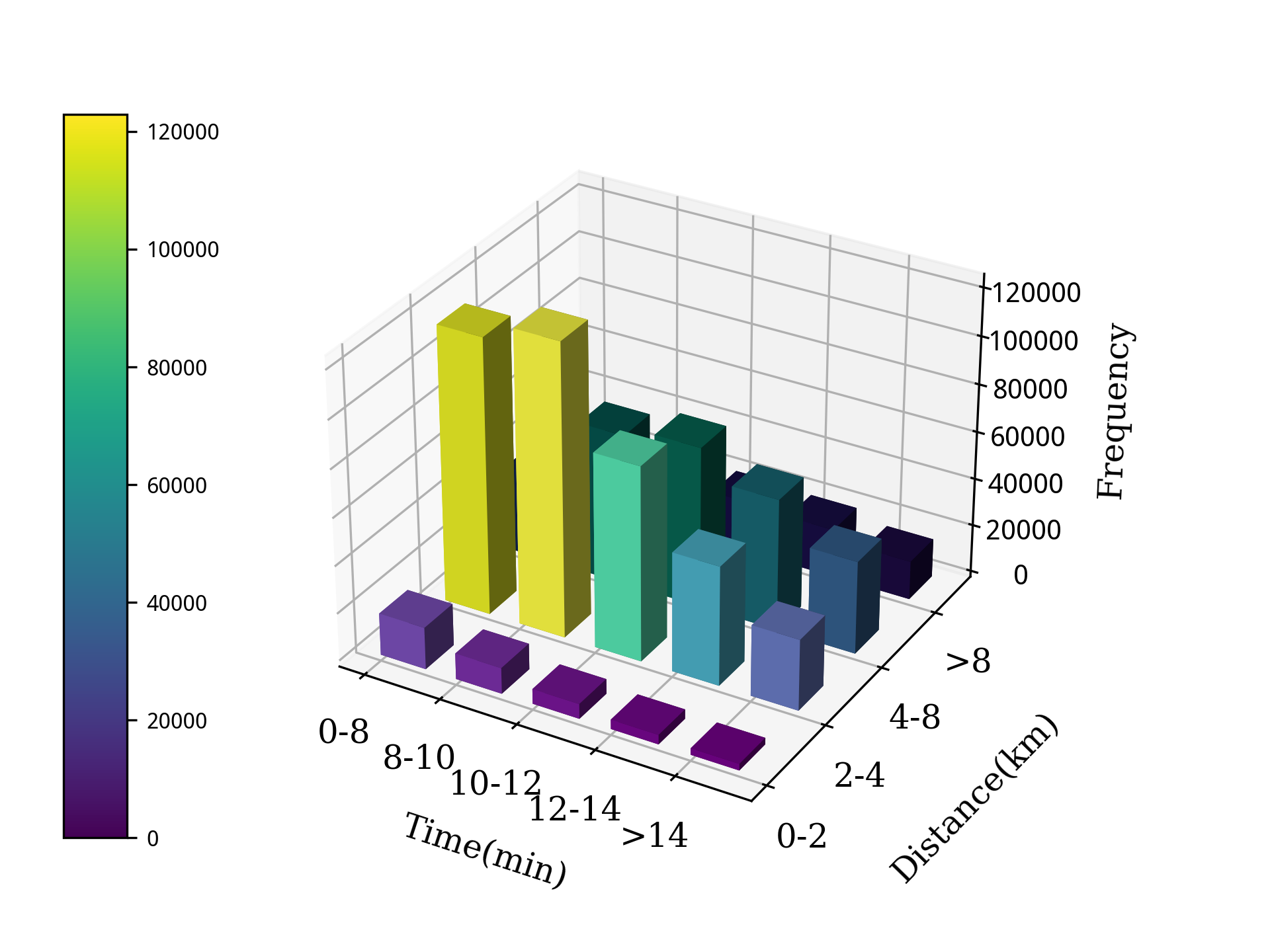}
  \label{fig:time_dist_chengdu_porto_2} } \caption{Frequency of time and distance on Chengdu and
  Porto datasets (CDF from 10\% to 80\%).}~\label{fig:time_dist_chengdu_porto}
  \end{figure}
  
  \begin{figure} \centering \subfloat[Chengdu
  dataset.]{\includegraphics[width=1.6in]{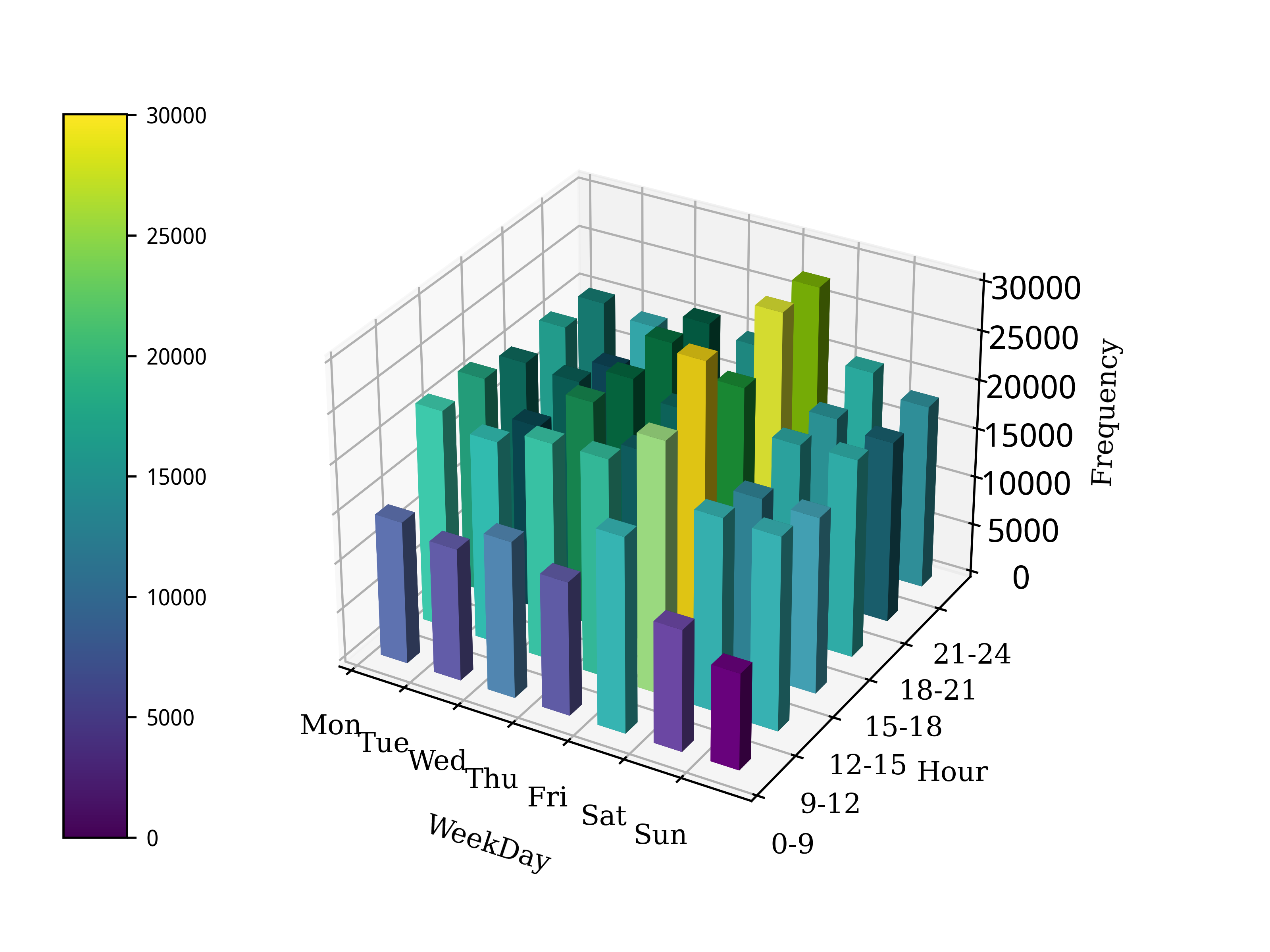}\label{fig:week_hour_chengdu_porto_1} } \hfil \subfloat[Porto
  dataset.]{\includegraphics[width=1.6in]{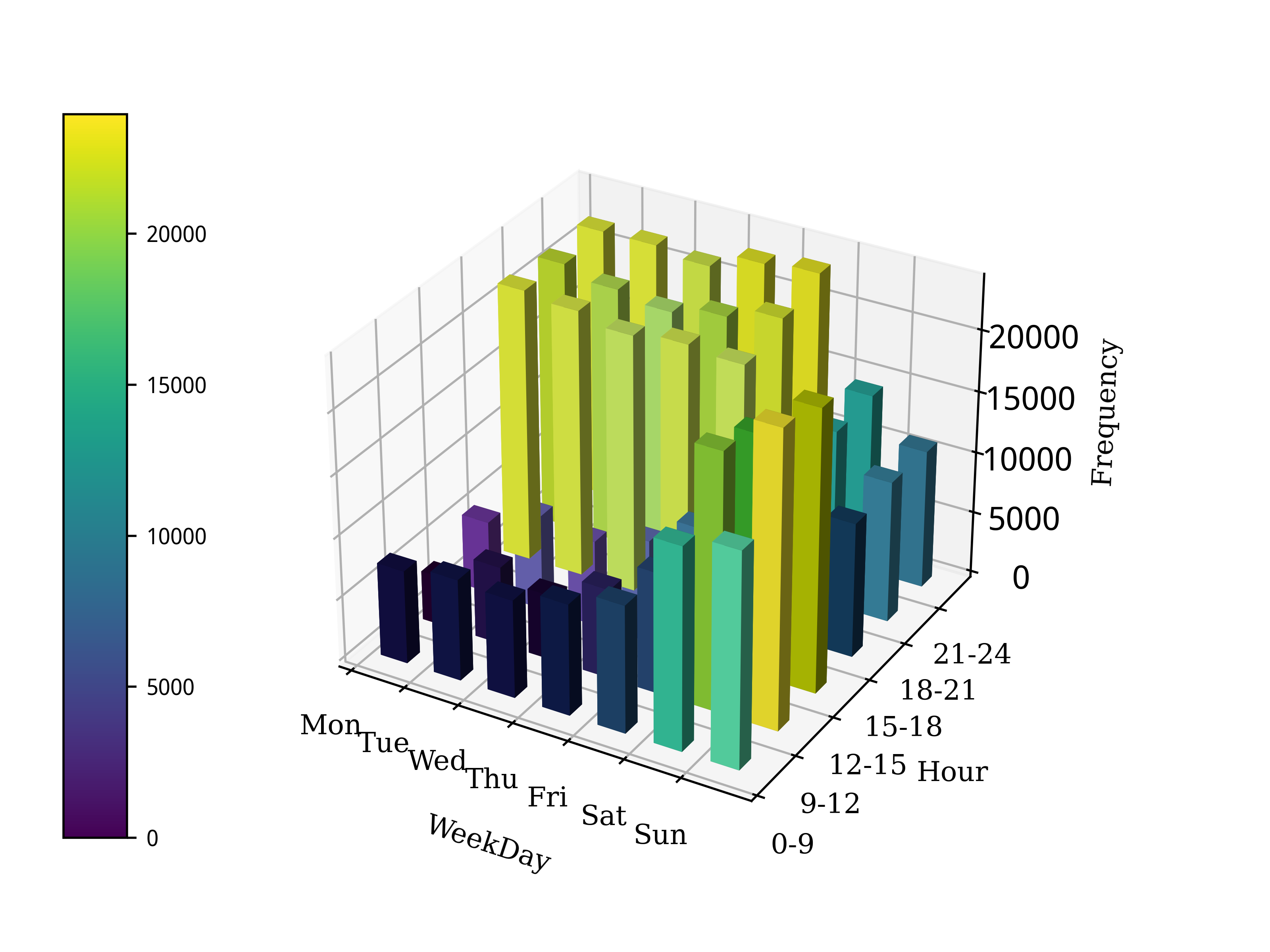}\label{fig:week_hour_chengdu_porto_2} } \caption{Frequency of day of week and hour on Chengdu
  and Porto datasets (CDF from 10\% to 80\%).}~\label{fig:week_hour_chengdu_porto}
  \end{figure}
We show the distribution of travel time of Chengdu and Porto datasets in Figure
~\ref{fig:cdf_travel_time}. Travel time on most trajectories (Cumulative
Distribution Function (CDF) from \textbf{10\%} to \textbf{80\%}) in Chengdu
dataset is between \textbf{315  seconds} and \textbf{1174 seconds} and that in
Porto is between \textbf{315 seconds} and \textbf{945 seconds}. Meanwhile, as
illustrated in Figure~\ref{fig:cdf_travel_distance}, travel distances on most
trajectories (CDF from \textbf{10\%} to \textbf{80\%}) in Chengdu dataset are
between \textbf{1.84 kilometers} and \textbf{8.14 kilometers} and that in Porto
are between \textbf{1.76 kilometers} and \textbf{7.32 kilometers}. We thereby
obtain available trajectories which fall into this range for our training phase
and by this approach the dirty data or the abnormal data are removed.
Furthermore, we also make more analysis on the Chengdu and Porto dataset to
observe their data frequencies in different temporal and spatial domains. As
illustrated in Figure~\ref{fig:time_dist_chengdu_porto}, the distributions of
data frequency of travel time and distance in Chengdu and Porto have some
similarities which indicates that people are more likely to take taxi trips in
the combination of travel time from 5 to 10 minutes and travel distance from 2
kilometers to 4 kilometers. However, Figure~\ref{fig:week_hour_chengdu_porto}
shows that the distributions of data frequency of the combination of the day of
week and hour in Chengdu and Porto are relatively different. People in Chengdu
seems to take more taxi trips from the afternoon to evening on Friday, while
people in Porto seems to take more taxi trips after 12:00 from Monday to Friday
or from morning to nightfall on Sunday. These different patterns of taking taxi
trips in multi-city scenarios may add difficulty to providing accurate travel
time estimations.

\section{Fine-grained Trajectory-based Travel Time Estimation Based on Deep Meta
Learning}~\label{sec:four}

\begin{figure}[h]
  \centering \includegraphics[width=8.5cm]{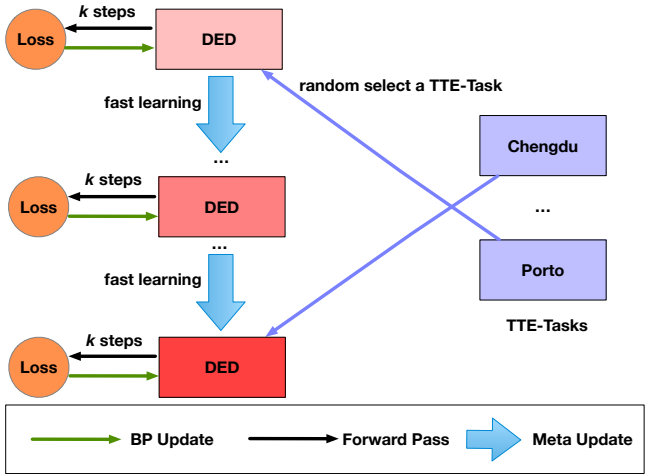} \caption{The
  framework of the proposed MetaTTE.\@ The DED consists of data preprocessing
  module and encoder-decoder network module, which is marked with color pink,
  light red and dark red indicating the three continuous states of parameter
  initialization.}\label{fig:metatte}
\end{figure}

\subsection{Overview}

Figure~\ref{fig:metatte} shows the framework of our proposed MetaTTE.\@ For each
iteration in the meta learning based training phase, a TTE-Task representing a
unique region is randomly selected at first and then is fed into the DED for $k$
times steps of regular optimization. At the end of each iteration, an adaptive
algorithm is adopted to change the initialization of parameters in DED in a fast
learning manner.

\begin{figure*}[h]
  \centering \includegraphics[width=12cm]{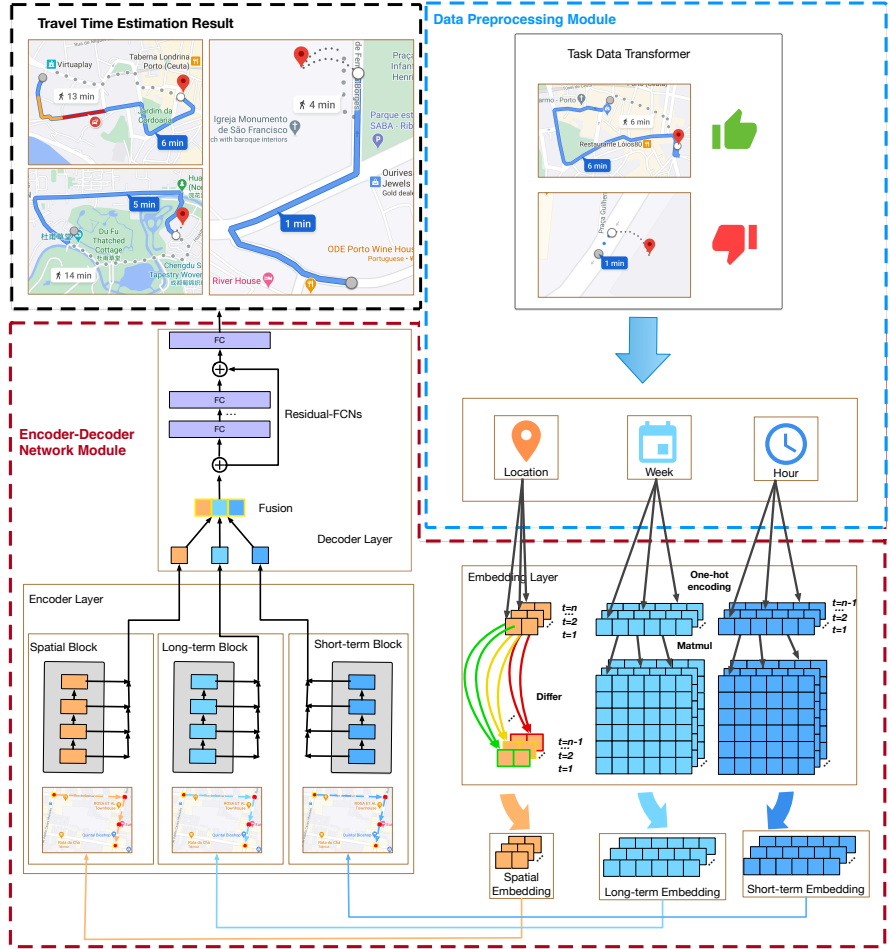} \caption{The
  architecture of DED.}~\label{fig:ded}
\end{figure*}

\subsection{DED}

As illustrated in Figure~\ref{fig:ded}, DED is composed of two components: the
\textbf{Data Preprocessing Module} and the \textbf{Encoder-Decoder Network
Module}. The Data Preprocessing Module is fed with the MetaTTE trajectory
firstly, to remove the biases of outliers in real-world datasets and transform
data into proper forms. Then the Encoder-Decoder Network Module is fed with the
preprocessed DED trajectory data, to learn, encode the spatial-temporal
representations and then decode the travel time estimations.

\subsubsection{Data Preprocessing Module}

The input pipelines are preprocessed in the Task Data Transformer and the
preprocessed data are fed into the Encoder-Decoder Network Module.

\textbf{Task Data Transformer.} Meta learning only uses small amount of examples
to train models for each iteration. To remove the biases of outliers in the
real-world datasets and quickly adapt to new possible tasks, we set up several
rules in the Data Preprocessing Module to dynamically control the input
pipelines for DED.

\begin{itemize}
\item \textbf{Rule 1:} Keep the most frequent trajectory data only. After the
deep insight into the analysis of our datasets, we decide only to keep
trajectories which satisfied all the requirements listed in Table
~\ref{tab:satis_datasets_rule1} to ignore the rarely appeared samples to avoid
possible biases when we train DED model using small amount of examples.

\item \textbf{Rule 2:} Keep the trajectory if and only if it contains at least
two different GPS points only. We regard that the isolated GPS point may be
caused by occasional positioning errors and it is also nonsense to navigate
users for the trajectory containing only one isolated GPS point.

\item \textbf{Rule 3:} Keep the trajectory if and only if the total travel time
is positive.

\end{itemize}

\begin{table}
  \caption{Satisfied requirements of Rule 1 for our datasets}\label{tab:satis_datasets_rule1} \begin{tabular}{l|l|l} \toprule Requirements
  & Chengdu & Porto\\ \midrule Travel Time No Less Than & 315 seconds & 315
  seconds\\ Travel Time No More Than & 1174 seconds & 945 seconds\\ Travel
  Distance No Less Than & 1.84 kilometers & 1.74 kilometers\\ Travel Distance No
  More Than & 8.14 kilometers & 7.32 kilometers\\ \bottomrule
\end{tabular}
\end{table}

\subsubsection{Encoder-Decoder Network Module}

The encoder-decoder network module mainly consists of three parts: the
\textbf{Embedding Layer}, the \textbf{Encoder Layer} and the \textbf{Decoder
Layer}. The preprocessed input data are firstly embedded into low dimensional
feature vectors in the Embedding layer and then different embeddings are encoded
separately in the Encoder layer through RNN. After that, the separate spatial
and temporal encoded feature representations are fused using Attention Mechanism
and then decoded using residual fully connected layers (Residual FCs) in the
\textbf{Decoder Layer}. And finally travel time estimations are made based on
the decoded feature representations.

\textbf{Embedding Layer.} In the embedding layer, we embed the the day of week
and hour which are represented with categorical values into low dimensional real
vectors, which allows neural network to conduct feature learning on our
real-world datasets. To better represent the temporal patterns for travel time
estimation, we choose a word embedding method mentioned in
~\cite{gal2016theoretically} which maps each categorical value $c\in [C]$ in
one-hot encoding to the embedding space (i.e. a real space) $\mathbb{R}^{E\times
1}$ by multiplying a trainable parameter matrix $W\in \mathbb{R}^{C\times E}$
with the initial values of all zeroes and then the parameter matrix is optimized
using the meta-learning based gradient descent algorithm
~\cite{ruder2016overview}.

In order to enhance the scalability of our model, all factors including the day
of week, hour are separately embedded to different channels, and the output for
each factor is formulated as: \begin{equation} e=\phi(x)\mathcal{W}^T
\end{equation} where $\phi(\cdot)$ represents the mapping function for one-hot
encoding, $x$ is the factor vector (i.e. the day of week and hour) and
$\mathcal{W}$ is the parameter matrix for feature learning. In this paper, the
differences of the GPS coordinates (i.e. $(\Delta p_1, \Delta p_2)$), the
embedding of the the day of week and the embedding of the hour are adopted to
represent the spatial embedding ($e_p$), the long-term embedding ($e_w$) and the
short-term embedding ($e_h$) respectively and are fed   into the encoder layer for
fine-grained feature learning.

\textbf{Encoder Layer.}  As mentioned in Section~\ref{sec:one}, it is significant to capture fine-grained spatial dependencies in daily scenarios for achieving satisfied travel time estimation. Some studies \cite{wang2018will} firstly apply convolutional layers to learn local-path features and then feeds local-path features into recurrent layer to further learn the entire-path features. However, since the size of kernels in the convolutional layer are fixed, the receptive field of local-path features learnt in this layer are also limited to fixed range. This prevents the model to learn fine-grained spatial features smaller than the size of kernels or larger than the size of kernels. To address this issue, we design fully RNN based blocks to encode the spatial embeddings, long-term embeddings, and short-term embeddings. Notice that DED supports user-specific blocks
for better scalability. It's simple to add or remove several blocks when needed 
since we employ feature learning in separate channels and this will not influence 
the other parts of DED.\@ Since state-of-the-art RNNs have been widely investigated,
we conduct extensive experiments on several common RNNs, including LSTM
~\cite{sak2014long}, GRU~\cite{chung2014empirical} and BiLSTM
~\cite{chiu2016named}. The update rule of each RNN is formulated as:
\begin{equation}
(\mathcal{C}^{t},\mathcal{Y}^{t})=RNN(e^t,\mathcal{C}^{t-1},\mathcal{Y}^{t-1})
\end{equation} where $\mathcal{C}^{t}$ is the updated hidden state and
$\mathcal{Y}^{t}$ is the updated output. Notice that there is no memory cell
$\mathcal{C}$ in GRU and we regard the $\mathcal{Y}$ as the hidden state
instead.

In this paper, we utilize the hidden states $C^t_{p},C^t_{w},C^t_{h}$ when $t=l$
as the learnt spatial-temporal features of spatial embedding, long-term
embedding and short-term embedding to transform variable length embeddings into
encoded fixed-length representations.

\textbf{Decoder Layer.} In the Decoder Layer, we first fuse the encoded
spatial-temporal features using \textbf{Attention Mechanism} and then decode the
fused spatial-temporal representations in \textbf{Residual FCs}.

\begin{itemize}
    \item \textbf{Attention mechanism.} To aggregate the encoded spatial-temporal features containing fine-grained spatial dependencies, a self-attention mechanism~\cite{bahdanau2014neural} is adopted to learn the importance of different
    dimensions in each type of feature and aggregate them to obtain fused
    spatial and temporal representations. We first concatenate all the
    spatial-temporal features (i.e. $\mathcal{C}_{p}, \mathcal{C}_{w},
    \mathcal{C}_{h}$) and then design a score function to automatically assign
    importance to different dimensions which is formulated as: 
    \begin{equation}
    \mathcal{S}=\max(0,((\mathcal{C}_{p} || \mathcal{C}_{w} ||
    \mathcal{C}_{h})\mathcal{W}+b)) 
  \end{equation} 
  where $\mathcal{C}_{p}, \mathcal{C}_{w}, \mathcal{C}_{h} \in \mathbb{R}^{N \times D}$ ($D$ is a
    hyper-parameter which indicates the output dimension of the embedding
    layer), $||$ represents the concatenation operation, $\mathcal{S}$
    represents the contribution score for different dimensions in each feature,
    $\mathcal{W} \in \mathbb{R} ^{N \times F \times F}$ ($F=3$) and $b$ are
    trainable parameters. Once  obtaining the scores for different dimensions,
    we normalize the score $\mathcal{S}^i$ for $i^{th}$ feature using the
    softmax function to obtain the attention values which can be formulated as:
    \begin{equation}
    a_i=Softmax(\mathcal{S}_i)=\frac{\exp(\mathcal{S}_i)}{\Sigma_i
    \exp(\mathcal{S}_i)} \end{equation} where $a_i$ is the attention value for
    the $i^{th}$ feature of each dimension. We calculate the fused spatial and
    temporal features for travel time estimation formulated as: 
    \begin{equation}
    \mathcal{C}_{f}=\Sigma_{i=1}^F a_i \odot {(\mathcal{C}_{p} || \mathcal{C}_{w}
    ||  \mathcal{C}_{h})}_i 
  \end{equation} 
  to obtain the fused spatial-temporal representation $\mathcal{C}_{f} \in \mathbb{R}^{N \times D}$.

\item \textbf{Residual FCs.} In practice, FCs are commonly used to decode the
spatial-temporal representations in higher level. Since the residual technique
can accelerate the training process  \cite{9078348}, we build residual blocks
with four FCs to decode the spatial-temporal representations and enhance the
performance without consuming much time. We formulate this process as:
\begin{equation} \mathcal{Y}_{estimation}=FC_{estimation}(FCs(\mathcal{C}_{f}) +
\mathcal{C}_{f}) \end{equation} where
$FC_i(\mathcal{Y})=\mathcal{Y}\mathcal{W}_i^T+b$, $\mathcal{W}_i$ is the
trainable parameter matrix for the $i^{th}$ fully connected layer and
\begin{equation} FCs(\mathcal{C}_{f}) = FC_1(FC_2(FC_3(FC_4(\mathcal{C}_{f})))).
\end{equation}
\end{itemize}

\subsection{Meta Learning based Optimization Algorithm}
\label{optimize}

\begin{figure}[h]
  \centering \includegraphics[width=6cm]{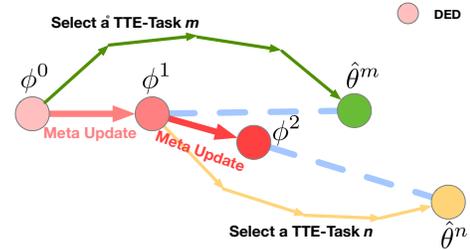} \caption{The
  schematic of the optimization algorithm for MetaTTE. Three circles marked with
  pink, red and dark red represent the continuous states of the initializations
  for trainable parameters in DED. Assume we firstly select TTE-Task $m$ and
  conduct 10 times gradient calculations to derive the parameters
  $\hat{\theta}^m$ (marked with green). Instead of arbitrarily updating the
  model using $\hat{\theta}^m$ at once, we conduct once meta-update which
  utilizes an adaptive algorithm to update the model parameters $\phi^0$ to
  $\phi^1$ based on $\hat{\theta}^m$. Then we select TTE-Task $n$ and conduct
  the same operations (marked with yellow) to update model parameters to
  $\phi^2$. This deferred update pattern helps MetaTTE pay more attention to the
  potential of providing accurate estimation further in the future.}
  \label{fig:meta}
\end{figure}

Inspired by Reptile \cite{nichol2018first}, which works by repeatedly sampling a
task, training on it and moving the initialization towards the trained
parameters on that task and has achieved good results on some well-established
benchmarks, we introduce the Reptile algorithm to optimize MetaTTE, which is
shown schematically in Figure \ref{fig:meta} and the pseudocode of the detailed
algorithm is shown in Algorithm \ref{algo:ours}.

The inputs of Algorithm \ref{algo:ours} are TTE-tasks $p_{\mathcal{T}}$, model
$\mathcal{M}$ which contain the hyperparameters $\lambda$ and the trainable
parameters $\theta$ and the loss function for task $\mathcal{T}_i$:
$\mathcal{L}_{\mathcal{T}_i}$. We first set a proper value for the maximum
iteration $\eta$ to declare the stopping criteria of the training phase. From
experiments, our MetaTTE can converge to the satisfied MAE (Mean Absolute Error)
metrics within 7000 iterations (i.e. $\eta=7000$). For each iteration in the
training phase, we firstly prepare datasets for training (Line 3). In the
training phase, we first save the current model parameters to $\theta_{1}$ (Line
4) and then sample $k$ batches of training data and conduct forward-backward
propagation using gradient descent based algorithm, Adam \cite{kingma2014adam},
to minimize the loss $\mathcal{L}_{\mathcal{T}_i}$ for optimization (Line
5$\sim$9). The model parameters after $k$ times of optimizations are saved in
$\theta_{2}$ (Line 10). Then we conduct the meta update on MetaTTE and calculate
$\theta_f$ (Line 11). Similar to Reptile, we develop the adaptive algorithm as a
linear learning rate scheduler \cite{gotmare2018closer} formulated as:
\begin{equation} \theta_f=\beta(1 - \frac{r}{\eta})(\theta_{2}-\theta_{1})
\end{equation} where $\beta$ is the step size for learning rate scheduling,
$\eta$ is the maximum iteration times, $r$ is the current training iteration,
$\theta_{1}$ and $\theta_{2}$ are the parameters before and after $k$ times of
training steps, respectively. At last, the model parameters are reset to
$\theta_f$ (Line 12) to accomplish optimization in this iteration.

\begin{algorithm}
\caption{Optimization algorithm for MetaTTE.} \label{algo:ours}
\SetNlSty{textbf}{}{:} \SetKwInOut{Input}{Input}  \SetKwInOut{Output}{Output}
\Input{TTE-Tasks $p_{\mathcal{T}}=\{\mathcal{T}_1,\mathcal{T}_2\}$, model
$\mathcal{M}$, loss function for task $\mathcal{T}_i$:
$\mathcal{L}_{\mathcal{T}_i}$.}  \Output{A partition of the bitmap} \BlankLine
randomly initialize $\theta$\; \For{$r$ in $[1, \eta)$}{  // \emph{multi-city
task selection} \; Randomly select a TTE-Task $\mathcal{T}_i$ from
$p_{\mathcal{T}}$\; Save current model parameters to $\theta_{1}$ \; \For{j in
$[1, k]$}{ $\mathcal{X}_{train}, \mathcal{Y}_{train}$ $\leftarrow$ Sample data
from $\mathcal{D}_{\mathcal{T}_i}^{train}$ \; $\hat{\mathcal{Y}}$ $\leftarrow$
$\mathcal{M}(\mathcal{X}_{train})$ \; Update model parameters using Adam on
$\mathcal{L}_{\mathcal{T}_i}$\; } Save current model parameters to $\theta_{2}$
\; Calculate model final parameters $\theta_f$\ using $\theta_{1},\theta_{2}$\;
Reset model parameters to $\theta_f$ \;
}
\end{algorithm}

\section{Experiments}
\label{sec:five}

In this section, we evaluate our proposed MetaTTE and compare it with six
baseline approaches based on Chengdu and Porto datasets.

\subsection{Baselines}

\noindent\textbf{AVG.~\cite{lan2019deepi2t}} This method calculates the average
speed with the travel time and the taxicab geometry in the training phase.
During the test phase, the estimation is calculated by averaging the historical
speeds of those with the same origin and destination.

\noindent\textbf{LR~\cite{lan2019deepi2t}.} This method trains the relation
between the taxicab geometry and travel time based on the locations of origin
and destination.

\noindent\textbf{GBM~\cite{lan2019deepi2t}.} This method utilizes the departure
time, the day of week, GPS coordinates and taxicab geometry to provide travel
time estimation using gradient boosting decision tree models.

\noindent\textbf{TEMP~\cite{wang2019simple}.} This method estimates the travel
time based on the average travel time using neighboring trips from large-scale
historical data.

\noindent\textbf{WDR~\cite{wang2018learning}.} This deep learning based method
extracts handcrafted features from raw trajectory data and utilizes information
extracted from road segments to provide travel time estimations.

\noindent\textbf{DeepTTE~\cite{wang2018will}.} This method learns spatial and
temporal feature representations from raw GPS trajectories and several external
factors using 1-D convolutional and LSTM networks.

\noindent\textbf{STNN~\cite{jindal2017unified}.} This method firstly
predicts the travel distance between an origin and a destination GPS coordinate,
and then combines this prediction with the time of the day to predict travel
time using fully connected neural networks.

\noindent\textbf{MURAT~\cite{li2018multi}.}  This method utilizes multi-task
representation learning to jointly learn the main task (i.e. travel time
estimation) and other auxiliary tasks (i.e. travel distance etc.) and enhances
the performance for travel time estimation.

\noindent\textbf{Nei-TTE~\cite{qiu2019nei}.} This method divides the entire
trajectory into multiple segments and captures features from road network
topology and speed interact using GRU for travel time estimation.

\noindent\textbf{MetaTTE-WA.} The variant of MetaTTE, which utilizes LSTM as the
RNN layer without attention based fusion.

\noindent\textbf{MetaTTE-WT.} The variant of MetaTTE, which utilizes LSTM as the
RNN layer without short-term and long-term embeddings.

\noindent\textbf{MetaTTE-LSTM.} The variant of MetaTTE, which utilizes LSTM as
the RNN layer.

\noindent\textbf{MetaTTE-BiLSTM.} The variant of MetaTTE, which utilizes BiLSTM
as the RNN layer.

\noindent\textbf{MetaTTE-GRU.} The variant of MetaTTE, which utilizes GRU as the
RNN layer.

\subsection{Experimental Settings} In this part, we first introduce the
configurations of our evaluation environment briefly and then describe the
hyperparameters in our MetaTTE.

\subsubsection{Configurations} We utilize the TensorFlow framework to implement,
train, validate and test our proposed MetaTTE. We conduct our evaluations on a node of
the Dawn supercomputer with the CPU (Intel E5-2680 2.4GHz x 28), RAM (64GB), GPU
(Tesla V100S 32GB), Operating System (Centos 7.4) and deep learning framework
(TensorFlow 2.3). During the test phase, we train 100 epochs for each baseline
using fine-tuned hyperparameters on Chengdu and Porto datasets respectively and
compare the results with our MetaTTE. Notice that conventional deep learning
methods optimize parameters on all batches of data for each epoch, while our
MetaTTE optimize parameters on small amount of data for each iteration, which is
much faster than the former.

\subsubsection{Hyperparameters} The hyperparameters for MetaTTE can be
categorized into two parts: (i) training hyperparameters which includes batch
size (32), step size $\beta (0.1)$, $k$ (10) batches of training data, maximum
iteration $\eta$ (7000); (ii) model hyperparameters which include the dimension
of embedding $D$ (64), the number of units in RNN $n_r$ (64), and the number of
units in residual FCs (1024, 512, 256, 64 respectively). Additionally, all the
trainable parameters in the model are initialized using Xavier initialization
method.

\begin{table*} \caption{Performance comparison of different baselines for travel
time estimation on Chengdu and Porto datasets. Notice that all metrics are
calculated based on travel time in seconds.} \centering
\begin{tabular}{l|c|c|c|c|c|c} \toprule 
 \multirow{2}{*}{Baselines} & \multicolumn{3}{c}{Chengdu} &
\multicolumn{3}{c}{Porto} \\ & MAE & MAPE (\%) & RMSE & MAE & MAPE (\%) & RMSE\\ 
\midrule 
AVG & 442.20 & 39.71 & 8443.60 & 182.64 & 26.66 & 1128.21 \\ 
LR & 516.23 & 49.09 & 1204.99 & 194.40 & 33.90 & 279.20\\
GBM & 454.50 & 41.67 & 1121.32 & \underline{148.53} & \underline{24.59} & \underline{209.07}\\
TEMP & \underline{334.60} & \underline{39.70} & \underline{761.05} & 174.44 & 28.73 & 260.81\\ 
\midrule 
WDR & 433.99 & 29.74 & 1024.92 & 164.04 & 22.84 & 244.41\\
DeepTTE & 413.09 & \underline{24.22} & \underline{926.04} & \underline{84.29} & \underline{14.79} & \underline{\textbf{90.29}}\\
STNN & 427.33 & 30.08 & 1011.88 & 226.30 & 35.44 & 331.75\\ 
MURAT & \underline{396.01} & 29.29 & 994.95  & 165.91 & 27.10 & 177.83\\ 
Nei-TTE  & 414.16 & 30.04 & 1038.71 & 106.30 & 15.23 & 183.03\\
\midrule 
MetaTTE-WT & 264.55 & 33.12 & 792.39 & 69.87 & 9.98 & 203.29\\ 
MetaTTE-WA & 258.10 & 27.16 & 774.87 & 68.16 & 9.84 & 204.64\\ 
MetaTTE-LSTM & 249.47 & 24.97 & 757.87 & 65.88 & 9.35 & 200.15\\
MetaTTE-BiLSTM & 254.47 & 25.55 & 766.45 & 67.20 & 9.59 & 202.42\\
MetaTTE-GRU & \underline{\textbf{236.38}} & \underline{\textbf{23.69}} & \textbf{\underline{745.11}} & \underline{\textbf{62.43}} & \underline{\textbf{8.83}} & \underline{196.78}\\ \bottomrule
\end{tabular}
\label{tab:baselines_results} \end{table*}

\subsubsection{Experimental Results}

In this part, we first compare the results among variants of MetaTTE and all of
six baselines. We firstly introduce the evaluation results using all the
baselines on overall datasets. Then in order to investigate the impacts of
different travel time and travel distances to estimation performance, we conduct
extensive experiments of baselines on two datasets. We then show our fine-tuning
results when investigating the impact of the hyperparameters. Lastly, we conduct
ablation studies on our MetaTTE.

\textbf{Comparing to all the baselines on Chengdu and Porto dataset.}
\label{results:all} Table \ref{tab:baselines_results} compares the performance of different baselines on two datasets. We can observe that: (1) deep learning based methods outperform other traditional time series methods in MAPE metric, which indicates the superior of its ability to learn dynamic temporal features and fine-grained spatial features for travel time estimation; (2) MetaTTE-GRU outperforms other baselines in MAE and MAPE metrics on two datasets. The reason for the
increase in RMSE metric on Porto dataset compared with DeepTTE may lie in: (i) instead of training two separate models for Chengdu and Porto like DeepTTE does,
MetaTTE-GRU trains only single model for both datasets of Chengdu and Porto
which may be slightly influenced by the volatile datasets; (ii) some outliers in datasets (rarely appeared samples with CDF below 10\% or above 80\%)  influence the RMSE metric since MetaTTE-GRU is fed with the raw trajectory data, while
DeepTTE is fed with the resampled trajectory data which reduces the influence of outliers. Since the MAPE metric is more important in real-applications for the fact that the user tolerance of the estimation gap varies according to total
travel time \cite{wang2018learning} and the objective of the applications is to satisfy users in majority, MetaTTE-GRU still outperforms other baselines; (3) The variants of MetaTTE significantly outperforms other baselines on Porto
dataset, which indicates the strong generalization ability of MetaTTE to
continuously provide accurate travel time estimation further in the future.

\begin{figure*}[!t] \centering \subfloat[MAE
metric.]{\includegraphics[width=2in]{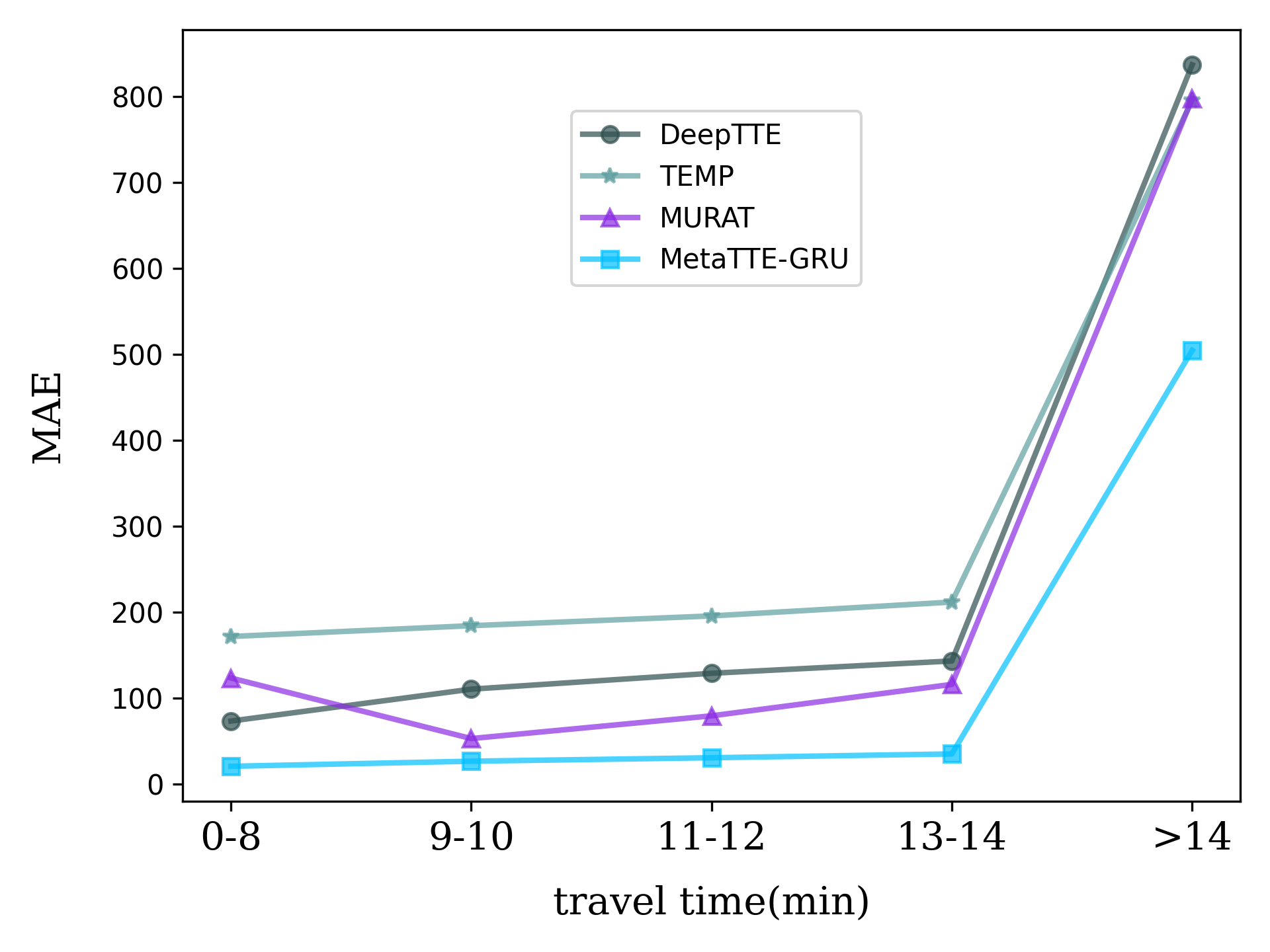}
\label{fig:results_chengdu_time_1} } \hfil \subfloat[MAPE
metric.]{\includegraphics[width=2in]{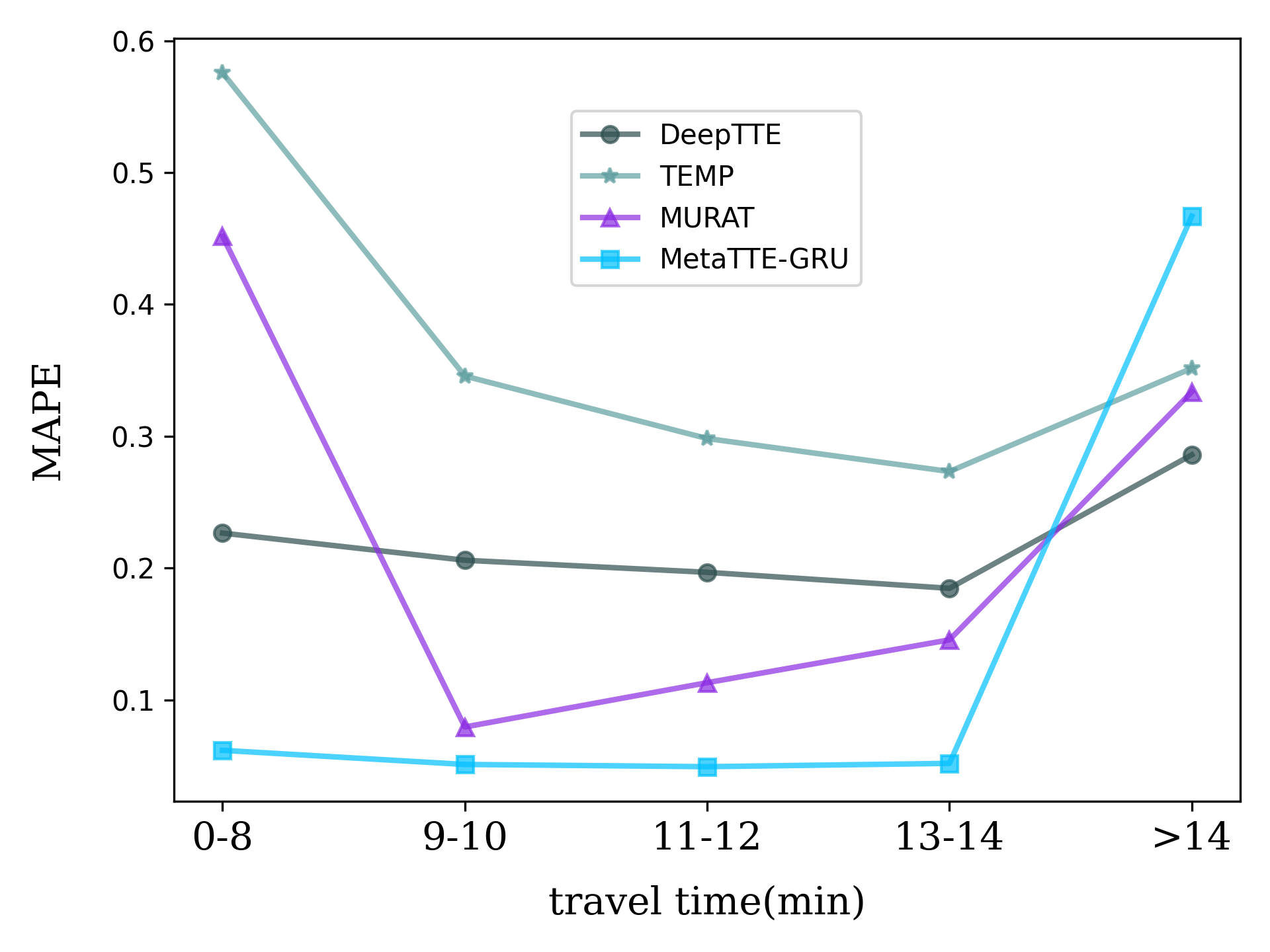}
\label{fig:results_chengdu_time_2} } \hfil \subfloat[RMSE
metric.]{\includegraphics[width=2in]{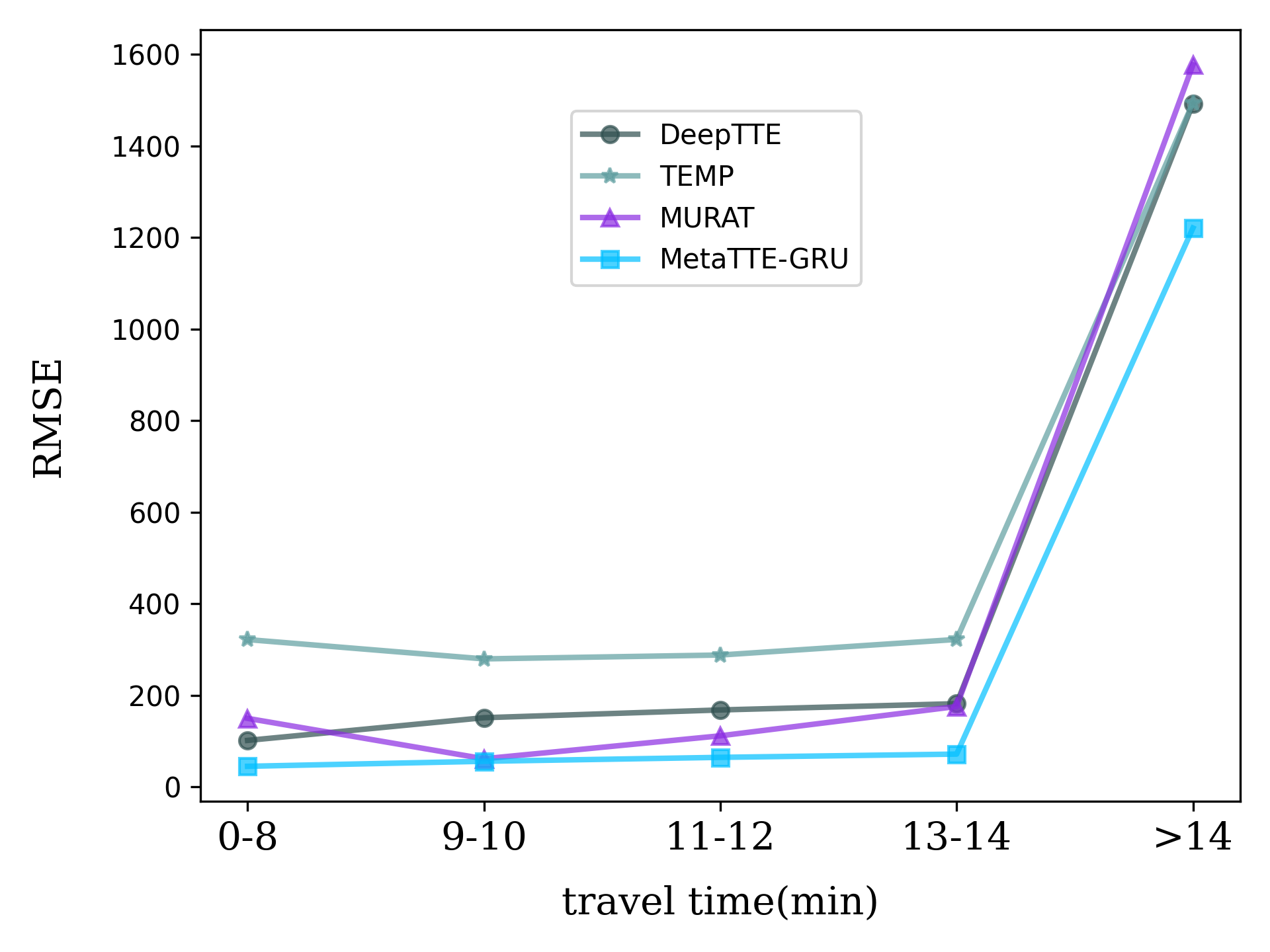}
\label{fig:results_chengdu_time_3} } \caption{MAE, MAPE and RMSE metric results on Chengdu
datasets of different travel time.} \label{fig:results_chengdu_time}
\end{figure*}

\begin{figure*}[!t] \centering \subfloat[MAE
metric.]{\includegraphics[width=2in]{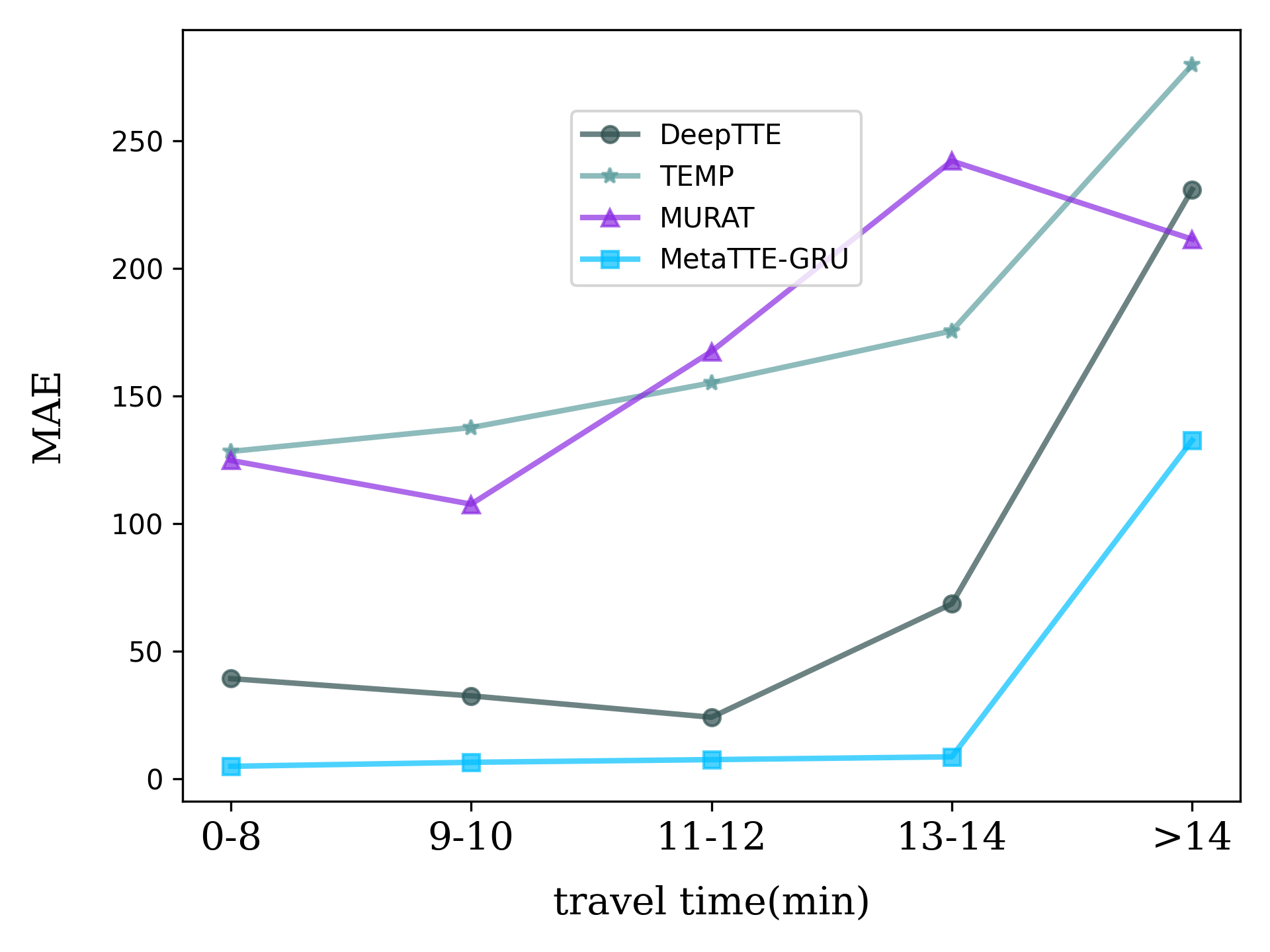}
\label{fig:results_porto_time_1} } \hfil \subfloat[MAPE
metric.]{\includegraphics[width=2in]{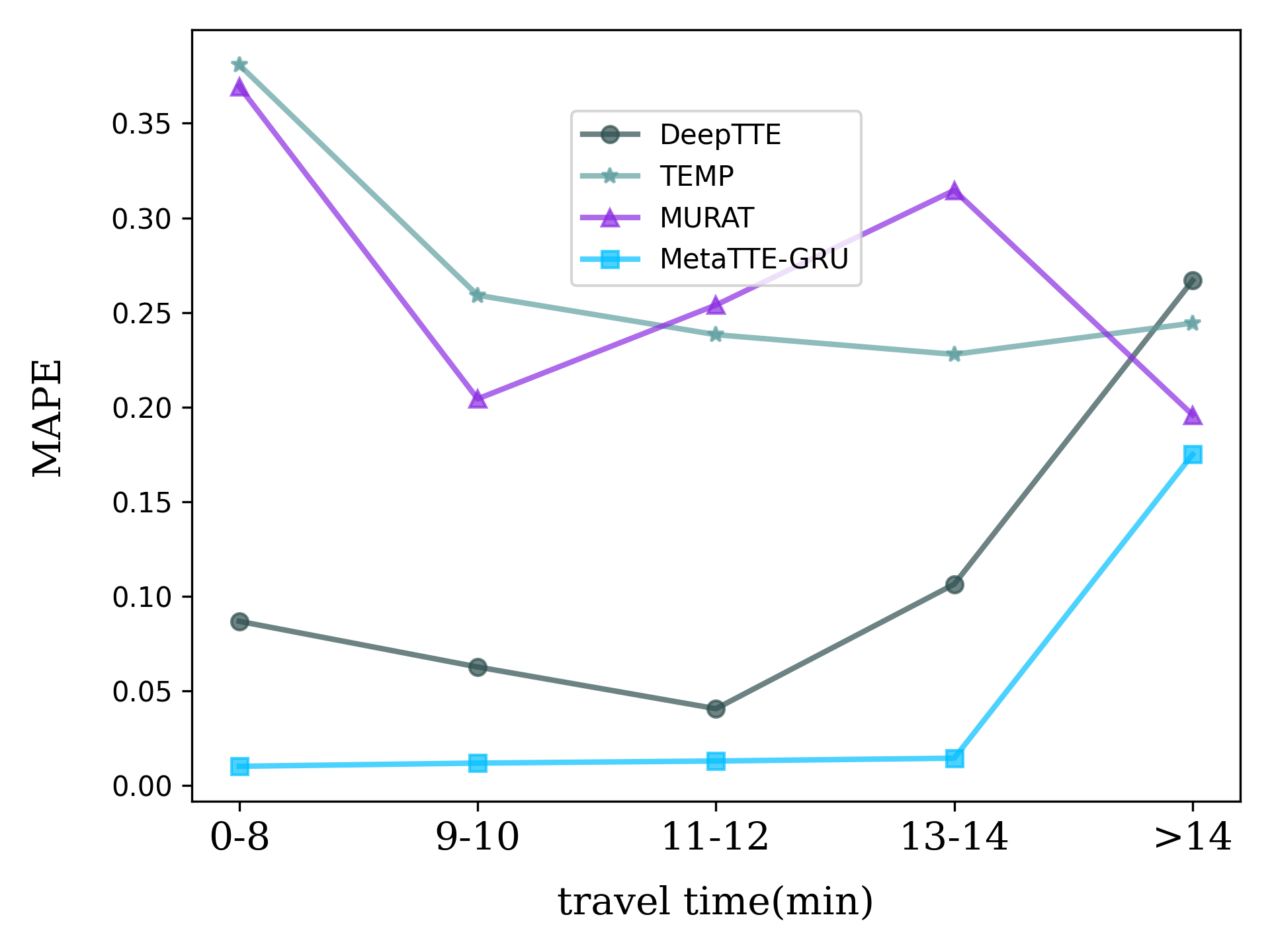}
\label{fig:results_porto_time_2} } \hfil \subfloat[RMSE
metric.]{\includegraphics[width=2in]{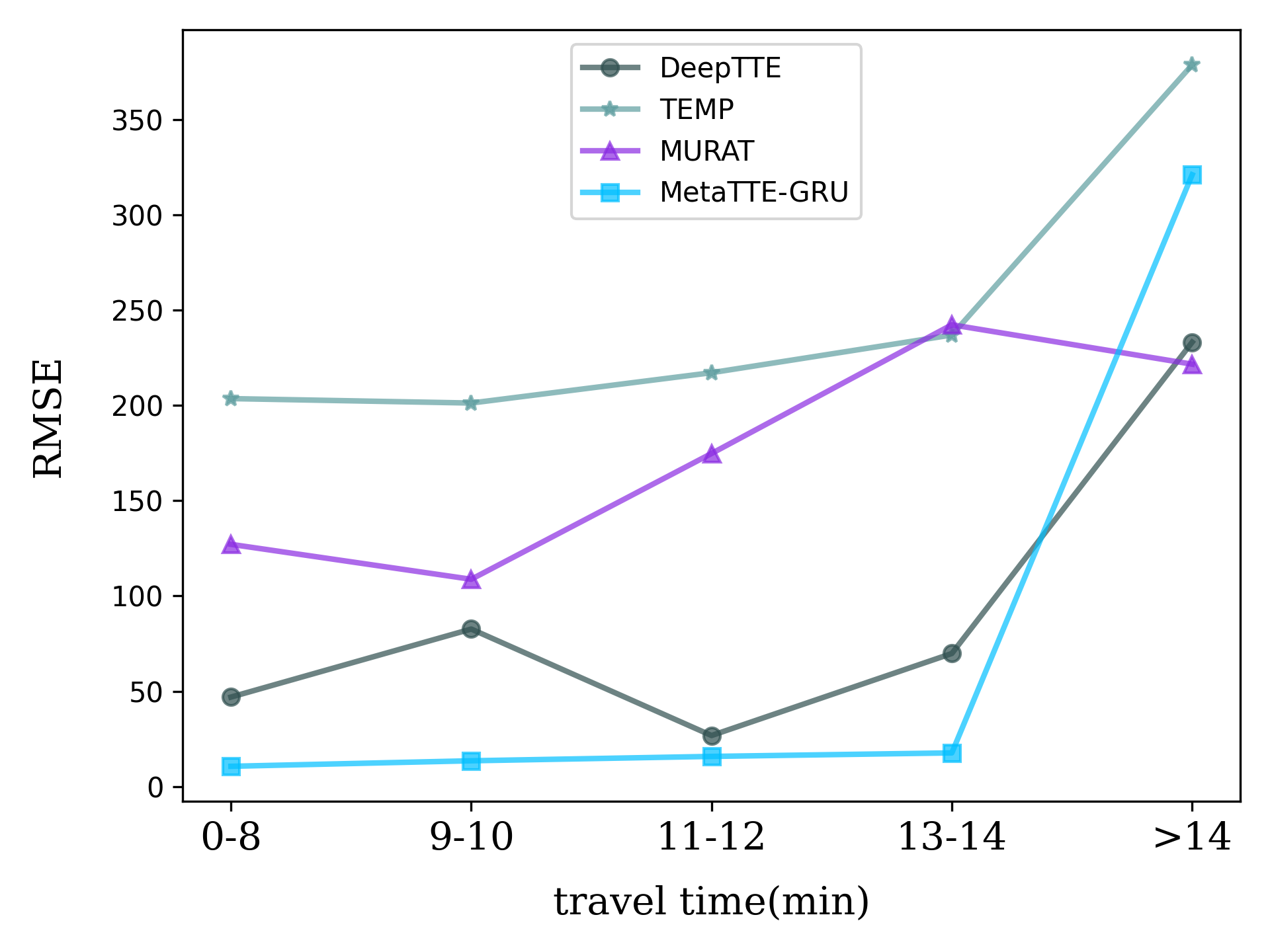}\label{fig:results_porto_time_3}
 } \caption{MAE, MAPE and RMSE metric results on Porto
datasets of different travel time.}\label{fig:results_porto_time} \end{figure*}

\textbf{Impact of different travel time.} In order to investigate the impact of
different travel time on MetaTTE comparing to selected baselines, we conduct
extensive experiments on different parts of the test datasets of both Chengdu
and Porto. Specially, we utilize MetaTTE-GRU which can outperform other
baselines using overall datasets in this study. As illustrated in Figure
\ref{fig:results_chengdu_time}, MetaTTE-GRU achieves the best performance in MAE
and RMSE metrics in different travel time except for the MAPE metric of the
travel time which is more than 14 minutes. According to the analysis of the data
in Section \ref{sec:three}, it is reasonable to regard these trajectory data as
the occasional events or the dirty data having be removed with the constraints
of Rule 1 in the data preprocessing module when training our model and this has
made this part of trajectory data the unseen dataset to our MetaTTE-GRU model.
Therefore, under such circumstances, MetaTTE-GRU can provide relatively better
estimation results in MAE and RMSE metrics which demonstrates the good
generalization ability for fault tolerance of MetaTTE-GRU to some extent. In a similar way, as shown in Figure \ref{fig:results_porto_time}, the performance of
MetaTTE-GRU on Porto dataset is better than other baselines in MAE and MAPE metrics.

\begin{figure*}[!t] \centering \subfloat[MAE
metric.]{\includegraphics[width=2in]{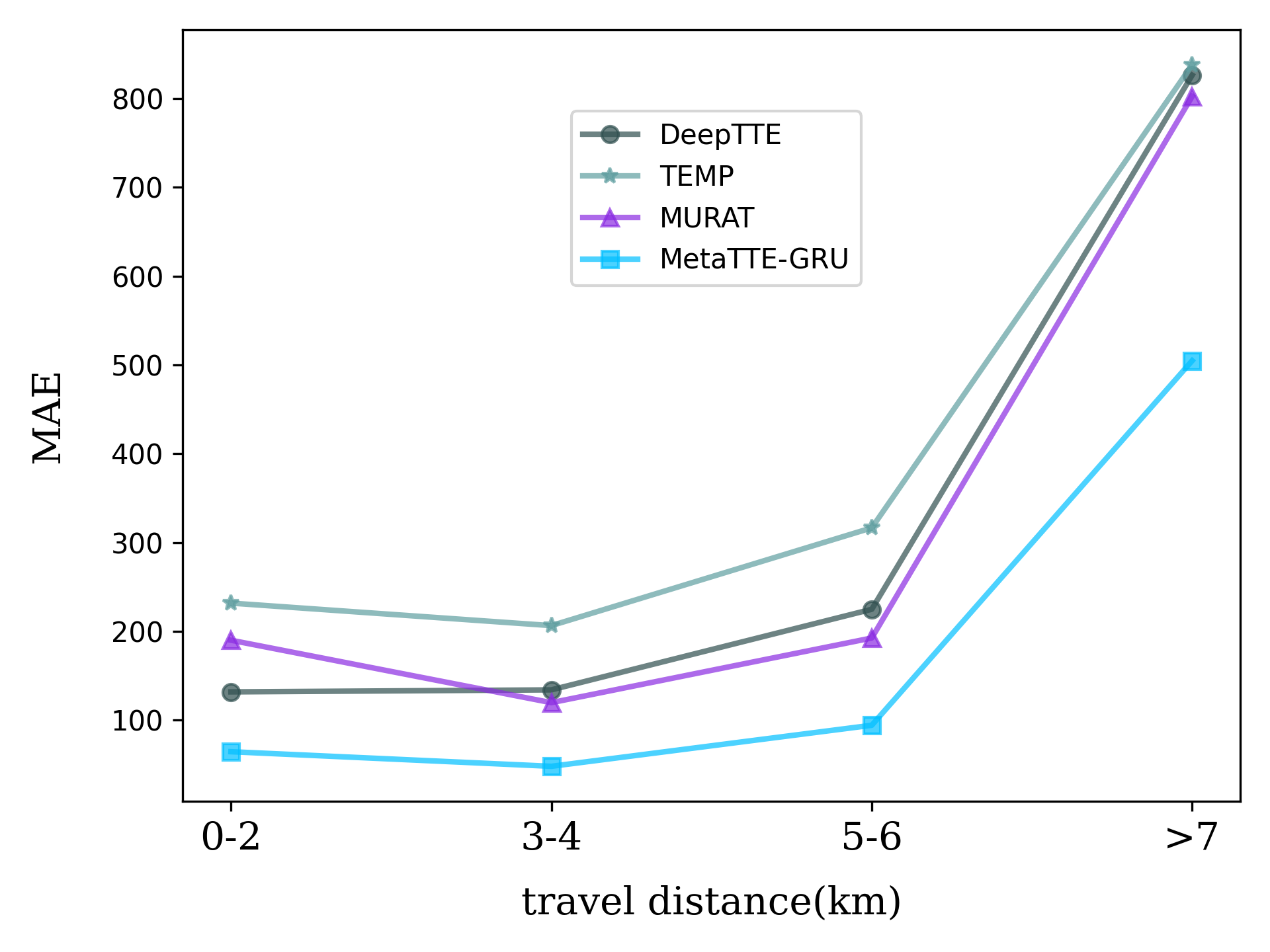}
\label{fig:results_chengdu_dist_1} } \hfil \subfloat[MAPE
metric.]{\includegraphics[width=2in]{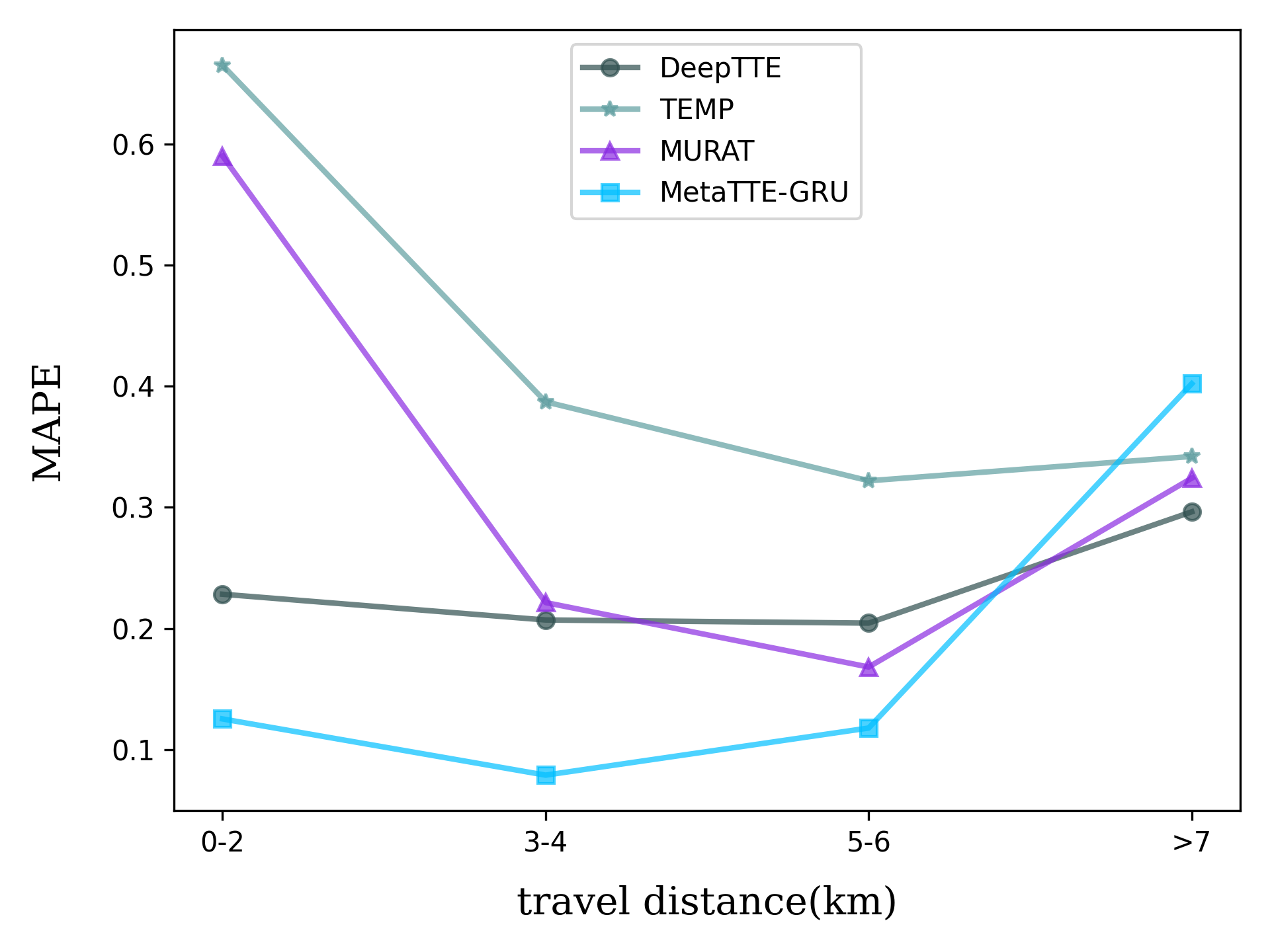}
\label{fig:results_chengdu_dist_2} } \hfil \subfloat[RMSE
metric.]{\includegraphics[width=2in]{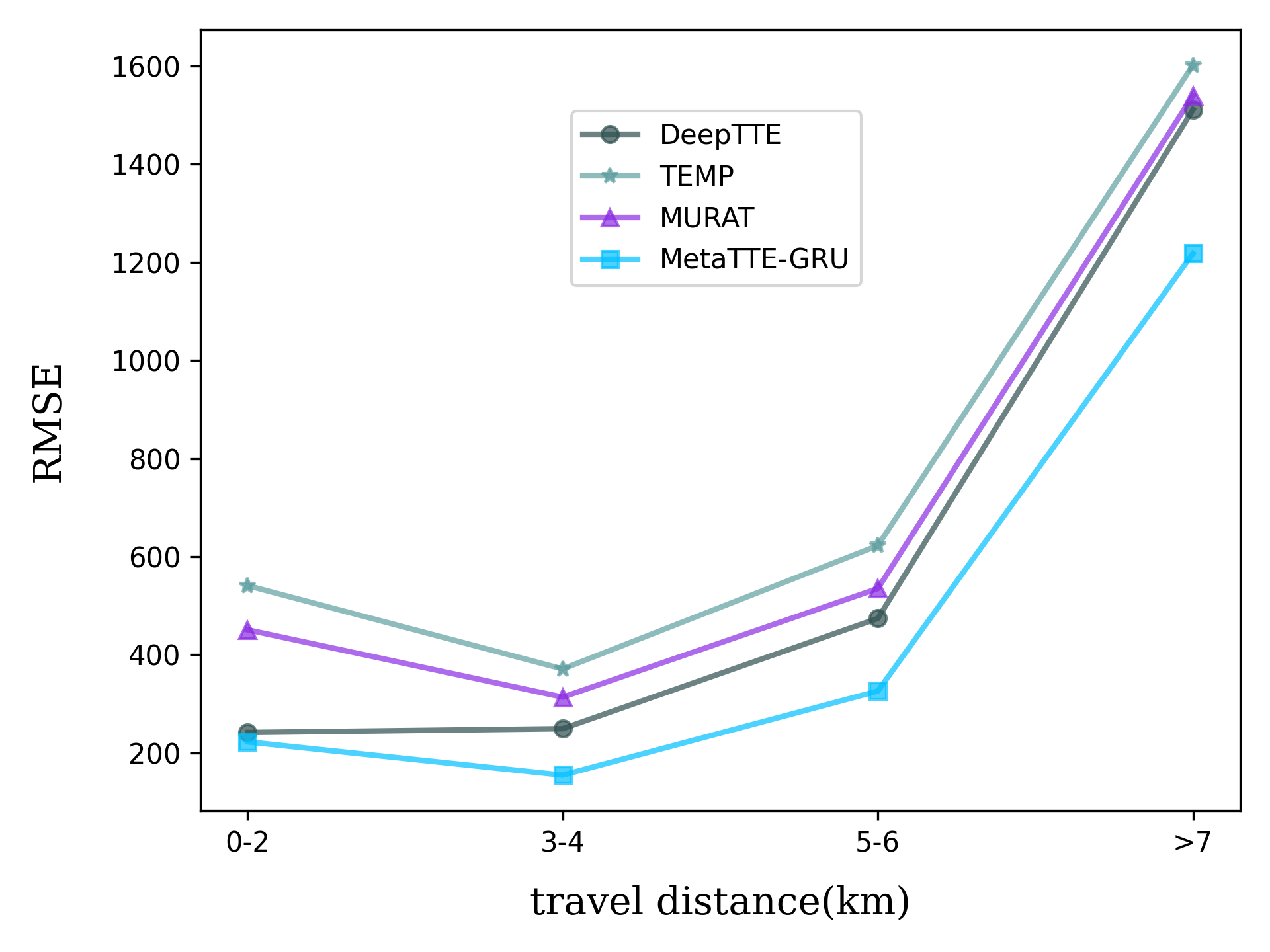}
\label{fig:results_chengdu_dist_3} } \caption{MAE, MAPE and RMSE metric results on Chengdu
datasets of different travel distances.}\label{fig:results_chengdu_dist}
\end{figure*}

\begin{figure*}[!t] \centering \subfloat[MAE
metric.]{\includegraphics[width=2in]{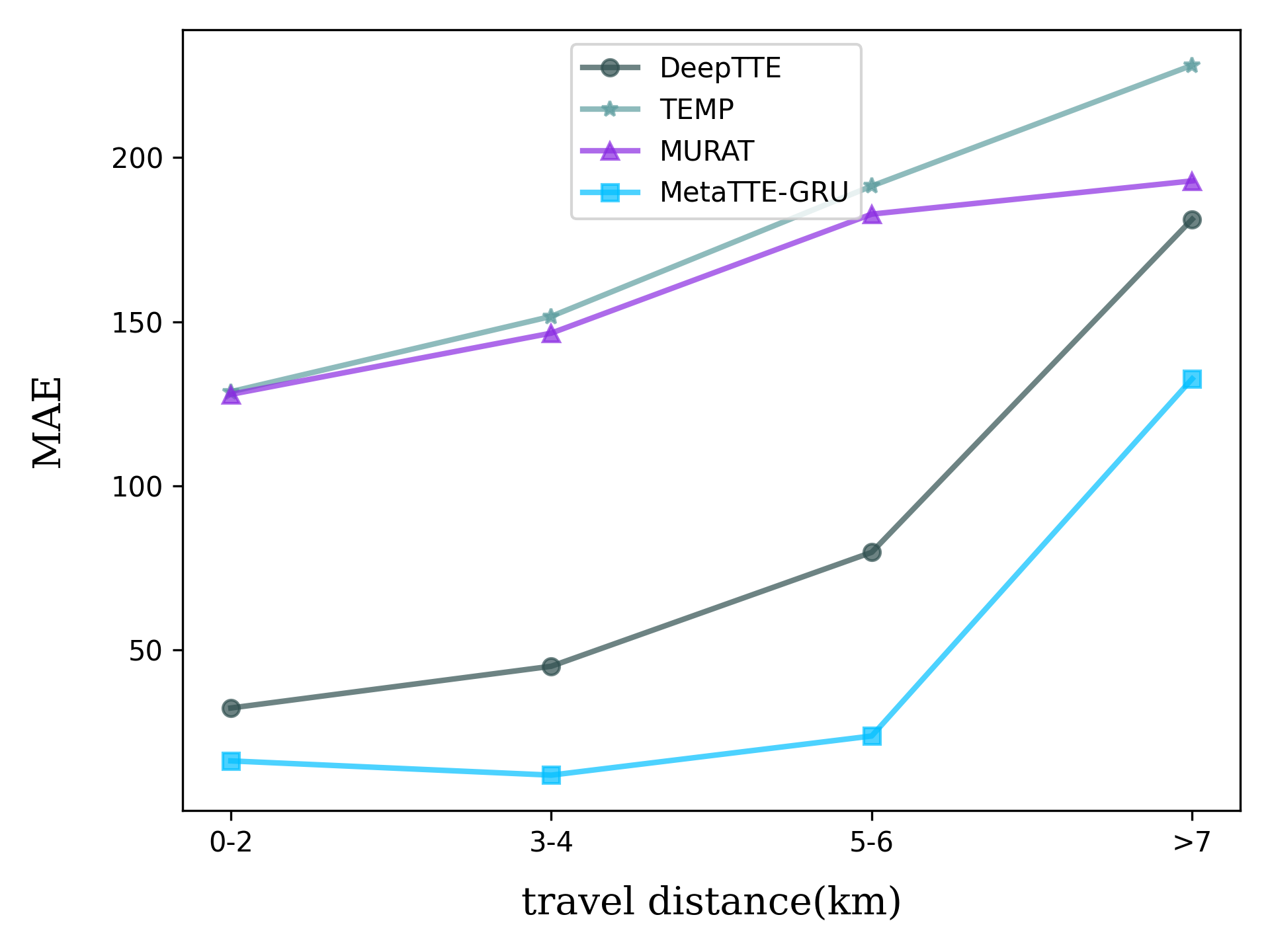}
\label{fig:results_porto_dist_1} } \hfil \subfloat[MAPE
metric.]{\includegraphics[width=2in]{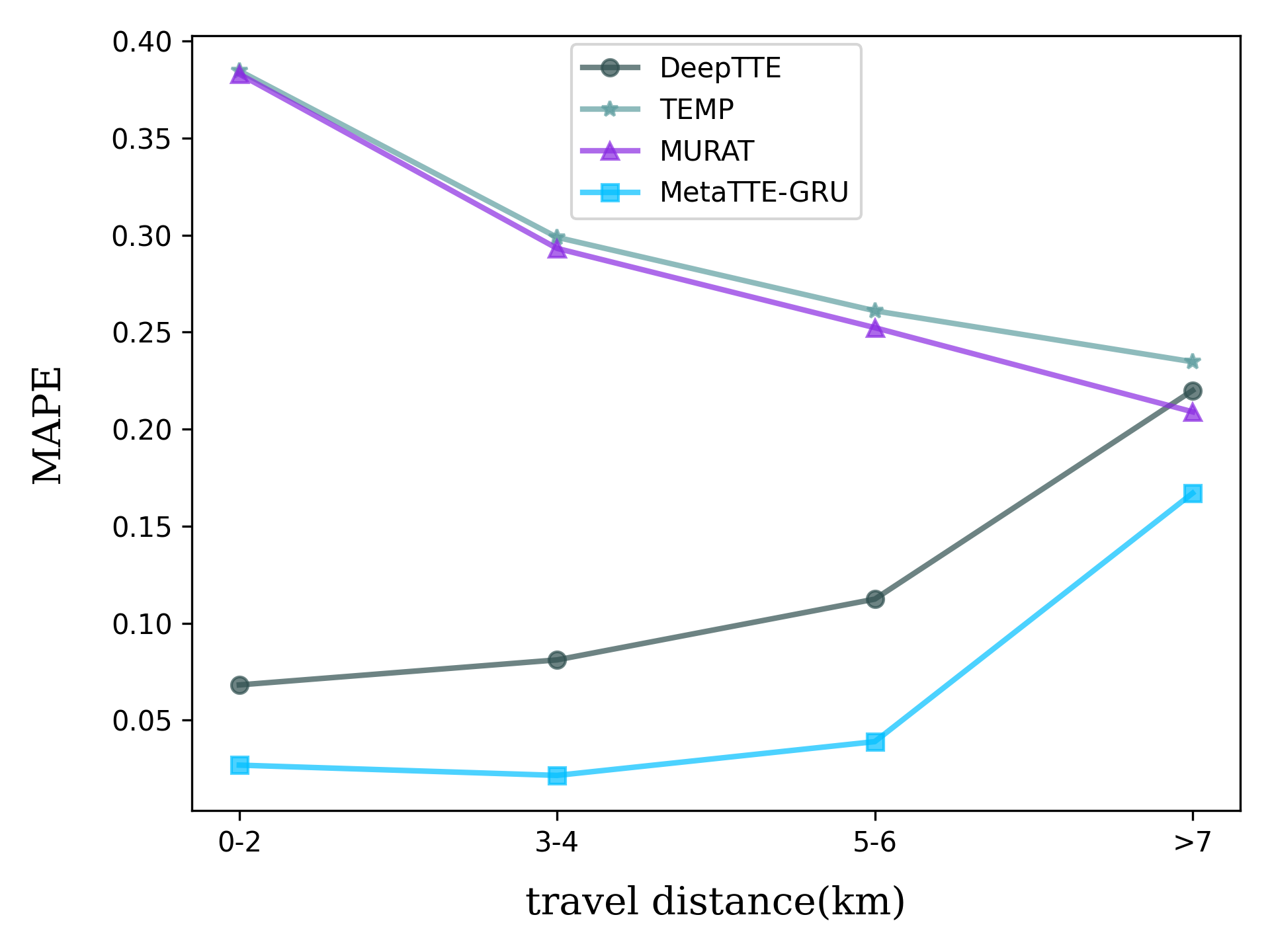}
\label{fig:results_porto_dist_2} } \hfil \subfloat[RMSE
metric.]{\includegraphics[width=2in]{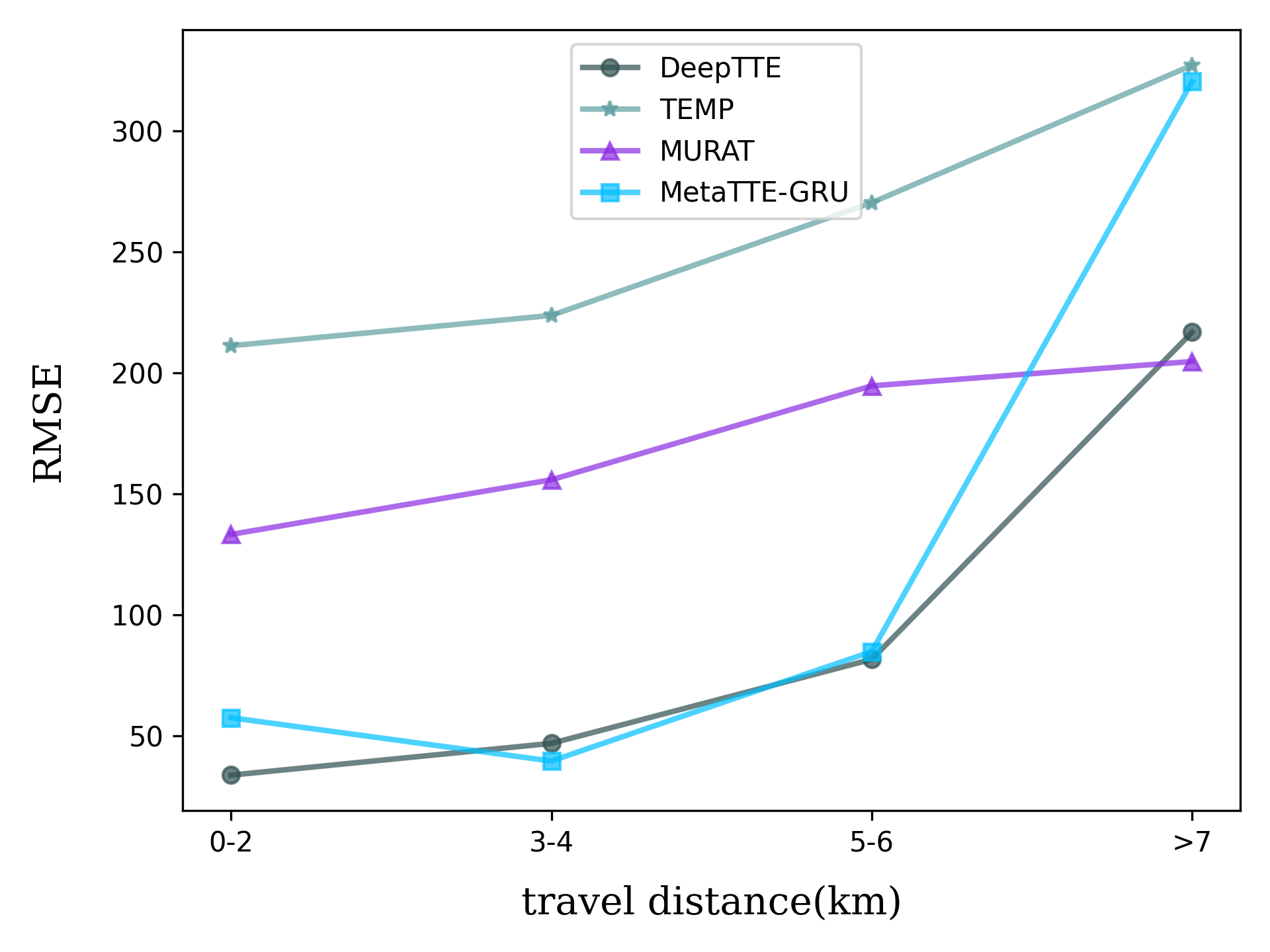}
\label{fig:results_porto_dist_3} } \caption{MAE, MAPE and RMSE metric results on Porto
datasets of different travel distances.}\label{fig:results_porto_dist}
\end{figure*}

\textbf{Impact of different travel distances.} We further investigate the impact
of different travel distances on MetaTTE-GRU and selected baselines. Similar
with the discussions made when investigating the impact of different travel
distances, Figure \ref{fig:results_chengdu_dist} and Figure
\ref{fig:results_porto_dist} show that MetaTTE-GRU achieves best performance in
most of the metrics (i.e. MAE and RMSE metrics) except for the MAPE metric on
occasional events which is higher than other baselines. Moreover, if we look
into the estimation results of TEMP, which is path-based methods, the MAPE
metric varies a lot when being tested on the most frequent travel distance data
and the rarest travel distance data which may indicate the poor fault tolerance
ability TEMP has. In a similar way, MURAT and DeepTTE may face the same pitfalls
while MetaTTE-GRU may not vary a lot in MAPE and RMSE metric when being tested
on those types of data and, especially, MetaTTE-GRU is trained on two datasets
with only single model which may be more likely biased than other baselines
which are trained as two individual models for two datasets.

\subsubsection{Impact of different hyperparameters.} In order to investigate the
impact of different hyperparameters and finetune our proposed MetaTTE, we
conduct experiments using different training and model hyperparameters.
Specially, we conduct experiments based on MetaTTE-LSTM which is also the
baseline we adopted in ablation studies. We then describe the results.

\begin{itemize}
    \item \textbf{Training hyperparameters.} The most important hyperparameter
    in our settings is the step size $\beta$ which schedules the learning rate
    for our meta-learning based optimization algorithm and thus we further
    investigate training hyperparameter and show the results in Table~\ref{tab:results_training_hyperparameters}. We can observe that the MAE,
    MAPE and RMSE metrics are decreasing from $\beta=0.05$ to $\beta=0.1$ while
    those are increasing from $\beta=0.1$ to $\beta=0.3$ which demonstrates that
    the best training hyperparameters are $\beta=0.1$ in general.
    \begin{table*}[!t] \caption{Evaluation results on different training
    hyperparameters.}~\label{tab:results_training_hyperparameters} \centering
    \begin{tabular}{l|c|c|c|c|c|c} \toprule \multirow{2}{*}{Hyperparameters} &
    \multicolumn{3}{c}{Chengdu} & \multicolumn{3}{c}{Porto}\\ & MAE & MAPE (\%)
    & RMSE & MAE & MAPE (\%) & RMSE\\ \midrule \textit{$\beta=0.05$} & 326.62 &
    35.88 & 815.20 & 86.26 & 12.76 & 215.29\\ \textbf{\textit{$\beta=0.1$}} &
    \textbf{249.47} & \textbf{24.97} & \textbf{757.87} & \textbf{65.88} &
    \textbf{9.35} & \textbf{200.15}\\ \textit{$\beta=0.3$} & 269.37 & 30.34 &
    797.49 & 71.14 & 10.55 & 210.62\\
     \bottomrule \end{tabular} \end{table*} \item \textbf{Model
    hyperparameters.} There are two key hyperparameters in our proposed MetaTTE,
    i.e., the dimension of embedding ($D$) and the number of units in RNN
    ($n_r$) and thus we finetune these two hyperparameters and present the
    experiment results in Table \ref{tab:results_model_hyperparameters}. We can
    observe that the MAE, MAPE and RMSE metrics are decreasing from
    $D=32,n_r=32$ to $D=64,n_r=64$ while those are increasing from $D=64,n_r=64$
    to $D=256,n_r=256$ which demonstrates that the best model hyperparameters
    are $D=64,n_r=64$ in general. \begin{table*}[!t] \caption{Evaluation results
    on different model hyperparameters.}
    \label{tab:results_model_hyperparameters} \centering
    \begin{tabular}{l|c|c|c|c|c|c} \toprule \multirow{2}{*}{Hyperparameters} &
    \multicolumn{3}{c}{Chengdu} & \multicolumn{3}{c}{Porto}\\ & MAE & MAPE (\%)
    & RMSE & MAE & MAPE (\%) & RMSE \\ \midrule \textit{$D=32,n_r=32$} & 302.07
    & 32.68 & 785.07 & 79.78 & 11.50 & 207.34\\ \textit{$D=64,n_r=64$} &
    \textbf{249.47} & \textbf{24.97} & \textbf{757.87} & \textbf{65.88} &
    \textbf{9.35} & \textbf{200.15}\\ \textit{$D=128,n_r=128$} & 295.84 & 33.08
    & 790.32 & 78.13 & 11.41 & 208.72\\ \textit{$D=256,n_r=256$} & 303.72 &
    33.74 & 795.29 & 80.21 & 11.74 & 210.04\\ \bottomrule \end{tabular}
    \end{table*}
\end{itemize}

\textbf{Ablation Studies.} In order to investigate the impact of components in
our proposed MetaTTE, we design several baselines, including MetaTTE-WT,
MetaTTE-WA and MetaTTE-LSTM to conduct ablation studies on Chengdu and Porto
datasets. \begin{itemize} \item \textbf{Impact of short-term and long-term
embeddings.} As illustrated in Table \ref{tab:baselines_results}, comparing
MetaTTE-WT with MetaTTE-WA, the MAE, MAPE and RMSE in Chengdu are decreased by
approximately 2.44\%, 18.00\% and 2.21\% and those in Porto are decreased by
approximately 2.45\%, 14.29\% and 0.66\%. Since the only difference between
MetaTTE-WA and MetaTTE-WT is that the former has short-term and long-term
embeddings while the latter doesn't, the short-term and long-term embeddings do
have positive impact on decreasing the prediction errors to some extent. \item
\textbf{Impact of attention mechanism based fusion component:} comparing
MetaTTE-WA with MetaTTE-LSTM, the MAE, MAPE and RMSE in Chengdu are decreased by
approximately 3.34\%, 8.06\% and 2.19\%, and those in Porto are decreased by
approximately 3.35\%, 4.98\% and 2.19\%. These results show that introducing
attention mechanism to the fusion component in MetaTTE can assign more fair
weights for spatial features, short-term features and long-term features which
can enhance the accuracy of travel time estimations.
\end{itemize}

\begin{table}
  \caption{Performance comparison of different baselines for travel time
  estimation on Didi Gaia dataset. Notice that all metrics are calculated
  based on travel time in seconds.} \centering \begin{tabular}{l|c|c|c}
  \toprule Baseline & MAE & MAPE (\%) & RMSE\\ \midrule AVG & \underline{135.40} & \underline{21.97} &
 451.31 \\ LR & 189.98 & 36.11 & 244.41\\ GBM & 194.42
  & 39.85 & 227.68\\ TEMP & 142.30 & 25.38 & \underline{220.41}\\ \midrule WDR &
  136.66 & 22.14 & 194.52\\ DeepTTE & 121.64 & 19.55 &
  \underline{\textbf{128.29}}\\ STNN & 173.67 & 30.90 & 231.08\\ MURAT &
  \underline{79.93} & \underline{11.25} & 138.15\\ Nei-TTE & 130.73 & 20.05 &
  192.26\\ \midrule MetaTTE-GRU & \underline{\textbf{59.76}} &
  \underline{\textbf{9.55}} & \underline{149.80}\\ \bottomrule
\end{tabular}

  \label{tab:didi_results}
\end{table}

\textbf{Comparing to baselines on fine-grained trajectory dataset.} To
further investigate the ability of baselines and MetaTTE to capture fine-grained
spatial dependencies, we introduce the Chengdu dataset in Didi Gaia dataset
~\cite{didi} and conduct experiments to evaluate the performance for travel time
estimation. The Didi Gaia dataset has a more regular and high sampling rate for
the trajectory data and contains 2,918,946 trips on the map of Chengdu over
Novermber, 2016. From Table~\ref{tab:didi_results}, we can observe: (1) deep
learning based methods outperform other conventional machine learning methods in MAPE metric,
which indicates the superior of their abilities to learn dynamic
spatial-temporal dependencies and fine-grained spatial features; (2) our
proposed MetaTTE-GRU outperforms other deep learning based methods in MAE and
MAPE metrics. The reason for the increase in RMSE metric compared with DeepTTE
is similar to that on Chengdu and Porto datasets which has been discussed in
Section~\ref{results:all}.

\textbf{Computation Complexity.} We present the training time of WDR,
DeepTTE, STNN, MURAT, Nei-TTE and variants for MetaTTE on Chengdu dataset in
Table~\ref{tab:time}. The training time of WDR, DeepTTE, STNN, MURAT, and NeiTTE
are calculated for 100 epochs and that of variants for MetaTTE are calculated
for 7000 iterations. We can observe that the training speed of variants of
MetaTTE are similar except for BiLSTM, which is much faster than DeepTTE.\@ WDR is
the most efficient but shows poor estimation performance. The training time of
STNN is similar to MetaTTE-WT but shows worse performance. These results have
demonstrated that MetaTTE balances the estimation performance and computation
burden, which is capable of achieving satisfied performance over time.

\begin{table}
  \caption{Time consumption for different baselines.}\label{tab:time}
  \centering \begin{tabular}{l|l} \toprule Baseline & Training Time (hrs)\\
  \midrule WDR & 1.3\\ DeepTTE & 160.7\\ STNN & 2.4\\ MURAT & 5.0\\ Nei-TTE &
  1.7\\ MetaTTE-WT & 2.5\\ MetaTTE-WA & 9.0\\ MetaTTE-LSTM & 10.0\\
  MetaTTE-BiLSTM & 20.0\\ MetaTTE-GRU & 9.5\\ \bottomrule
\end{tabular}
\end{table}

\section{Related work}\label{sec:six}

Existing works on travel time estimation generally fall into two categories:
\textbf{segment based methods} and \textbf{end-to-end methods}.

The segment-based methods~\cite{jenelius2013travel,
asif2013spatiotemporal,yang2013travel,lv2014traffic, ma2015long} sum up the
estimation results of individual road segments along the whole path to acquire
the travel time. Since most of the prior studies do not consider the
correlations or interactions among road segments, local errors along the road
segments will accumulate and thus lead to large errors.

The end-to-end methods mainly fall into two categories: similar paths based
methods~\cite{luo2013finding,rahmani2013route,wang2019simple} and deep learning
based methods
~\cite{zhang2018deeptravel,wang2018learning,qiu2019nei,fu2019deepist,fang2020constgat,zhang2020real,wang2018will,li2019learning,lan2019deepi2t,r1_1}.
The former tends to find the similar paths or neighbors of the query path to
estimate travel time, which cannot obtain good results due to data sparsity and
fluctuation issues. The latter utilizes large amount of historical traffic data
to build their models. DeepTravel~\cite{zhang2018deeptravel} firstly partitions
the whole road network into $N \times N$ grids and then extracts dual-term
features to establish a BiLSTM networks for travel time estimations. WDR
~\cite{wang2018learning} proposes a generalized model which is composed of
multiple fully-connected layers and a recurrent model, to learn the features in
spatial, temporal, road network and personalized information for travel time
estimation. ConSTGAT~\cite{fang2020constgat} firstly extracts features from
historical traffic conditions and background information, and then utilizes a
graph attention mechanism to capture the spatial-temporal relations of traffic
conditions. It provides the travel time of both the links in the route and the
whole route using a multi-task mechanism. PathRank~\cite{r1_1} is a
context-aware multi-task learning framework which estimates the ranking scores
for candidate routing paths as the main task, along with auxiliary tasks
including travel time estimation. PathInfoMax~\cite{r1_2} is an unsupervised
path representation learning based framework with curriculum negative sampling,
which produces path representations based on historical trajectory data and road
network information without task-specific labels. However, the performance of
these methods heavily rely on road network data which required extensive map
matching computation and is influenced by the time-varying circumstances. To
tackle these problems, some studies utilize traffic data without road networks
after careful preprocessing (resampling to relatively fixed patterns
~\cite{wang2018will} or morphological layouts with traffic states
~\cite{lan2019deepi2t} etc.) to learn the spatial and temporal features for
travel time estimation. However, these methods rely on careful preprocessing on
large amount of traffic data to achieve satisfied performance and are difficult
to continuously provide accurate travel time estimations over time.

\section{Conclusion}\label{sec:seven}

We investigate the fine-grained trajectory-based travel time estimation problem
for multi-city scenarios. We construct two TTE-Tasks from two real-world
datasets for training. We propose a novel meta learning based framework,
MetaTTE, which opens up new opportunities to continously provide accurate travel
time estimations over time, especially the potential to achieve consistent
performance when traffic conditions and road network change over time in the
future. We propose a deep neural network model, DED, in MetaTTE which consists
of Data preprocessing module and Encoder-Decoder network module to embed the
spatial and temporal attributes into spatial, short-term, and long-term
embeddings using RNN, fuse high level features using attention mechanism and
capture fine-grained spatial and temporal representations for accurate travel
time estimation. Extensive experiments based on two real-world datasets verify
the superior performance of MetaTTE over six baselines. We hope our framework
could be used for travel time estimation in real ITS applications. It is also
significant to utilize meta learning techniques in MetaTTE, to continuously
provide accurate travel time estimation over time.

\bibliographystyle{IEEEtran}
\bibliography{IEEEabrv, IEEEfull}

\begin{IEEEbiography}[{\includegraphics[width=1in,height=1.25in,clip,keepaspectratio]{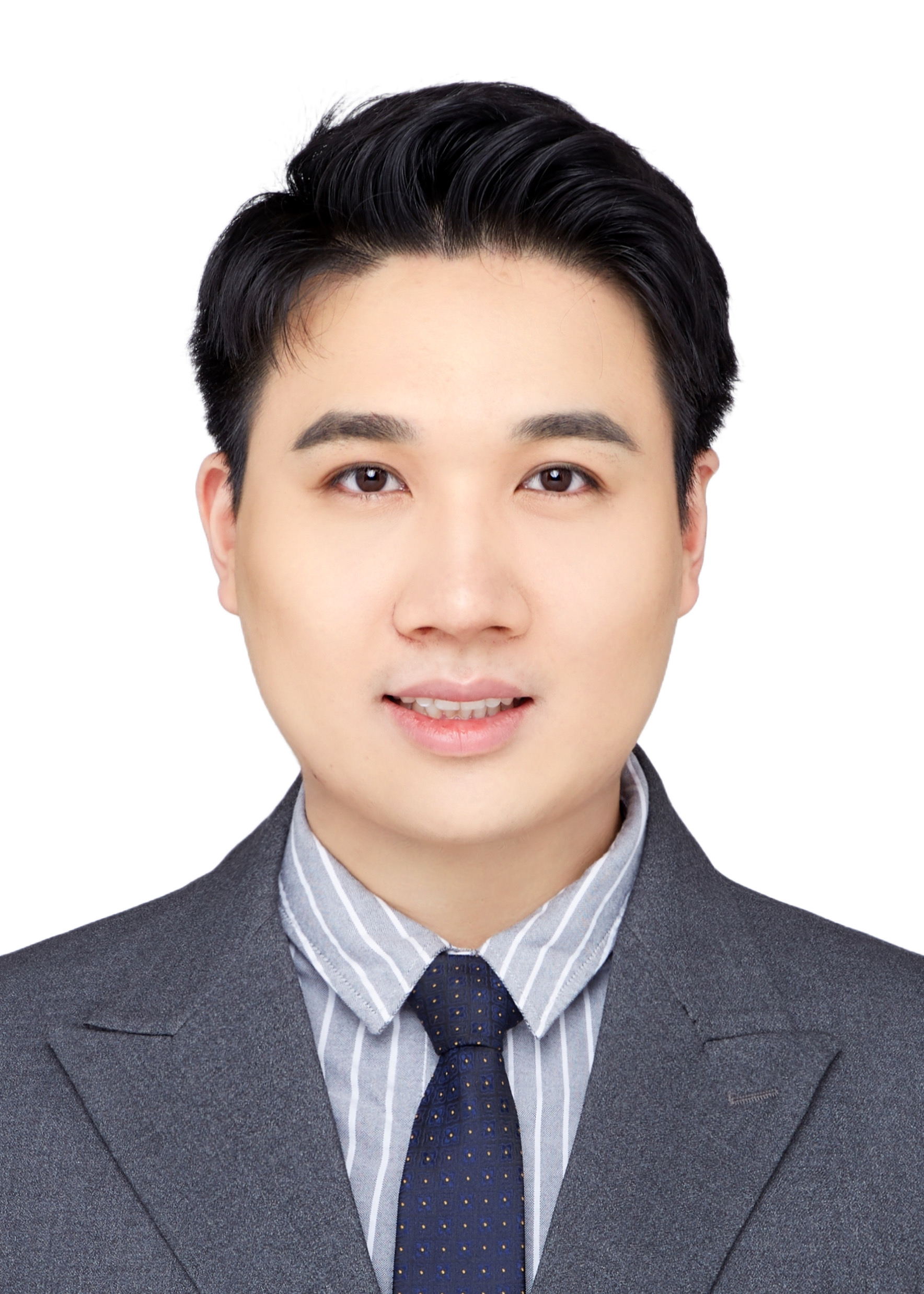}}]{Chenxing
Wang} is currently pursuing the Ph.D. degree with the School of Computer
Science (National Pilot Software Engineering School), Beijing University of
Posts and Telecommunications, Beijing, China. His current main interests
include spatial-temporal data mining, travel time estimation, traffic flow
prediction and transportation mode detection using deep learning techniques.
\end{IEEEbiography} \vskip 0pt plus -1fil

\begin{IEEEbiography}[{\includegraphics[width=1in,height=1.25in,clip,keepaspectratio]{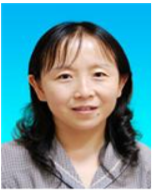}}]{Fang Zhao} received the B.S degree in the School of Computer Science and Technology from Huazhong University of Science and Technology, Wuhan, China in 1990, M.S and Ph.D. degrees in Computer Science and Technology from Beijing University of Posts and Telecommunication Beijing China in 2004 and 2009, respectively. She is currently Professor in School of Software Engineering Beijing University of Posts and Telecommunication. Her research interests include mobile computing, location-based services and computer networks.
\end{IEEEbiography} \vskip 0pt plus -1fil

\begin{IEEEbiography}[{\includegraphics[width=1in,height=1.25in,clip,keepaspectratio]{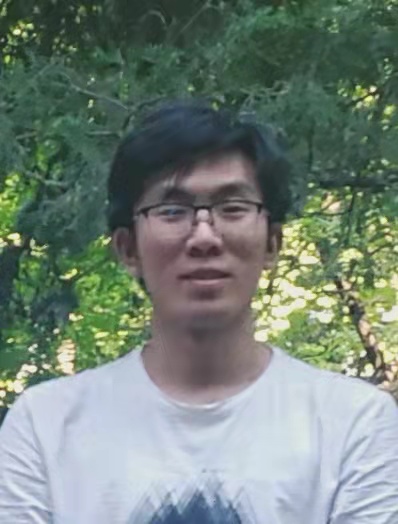}}]{Haichao
Zhang} received the B.E degree in the Department of Software Engineering from
Beijing University of Posts and Telecommunications, Beijing, China. He is
currently pursuing the M.S with the School of Software Engineering, Beijing
University of Posts and Telecommunications, Beijing, China. His current main
interests include spatial-temporal data mining, travel time estimation methods
using deep learning techniques. \end{IEEEbiography} \vskip 0pt plus -1fil

\begin{IEEEbiography}[{\includegraphics[width=1in,height=1.25in,clip,keepaspectratio]{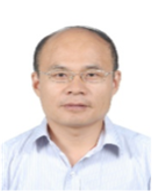}}]{Haiyong
Luo} received the B.S degree in the Department of Electronics and Information
Engineering from Huazhong University of Science and Technology, Wuhan, China
in 1989, M.S degree in School of Information and Communication Engineering
from the Beijing University of Posts and Telecommunication China in 2002, and
Ph.D. degree in Computer Science from the University of Chines Academy of
Sciences, Beijing China in 2008. Currently he is Associate Professor at the
Institute of Computer Technology, Chinese Academy of Science (ICT-CAS) China.
His main research interests are Location-based Services, Pervasive Computing,
Mobile Computing, and Internet of Things. \end{IEEEbiography} \vskip 0pt plus
-1fil

\begin{IEEEbiography}[{\includegraphics[width=1in,height=1.25in,clip,keepaspectratio]{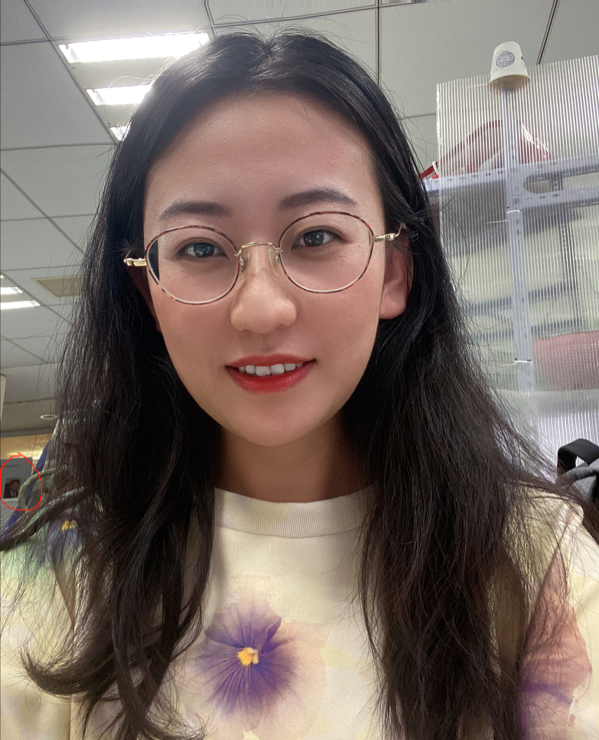}}]{Yanjun
Qin} is currently pursuing the Ph.D. degree with the School of Computer
Science (National Pilot Software Engineering School), Beijing University of
Posts and Telecommunications, China. Her current main interests include
location-based services, pervasive computing, convolution neural networks and
machine learning. And mainly engaged in traffic pattern recognition related
project research and implementation. \end{IEEEbiography} \vskip 0pt plus -1fil

\begin{IEEEbiography}[{\includegraphics[width=1in,height=1.25in,clip,keepaspectratio]{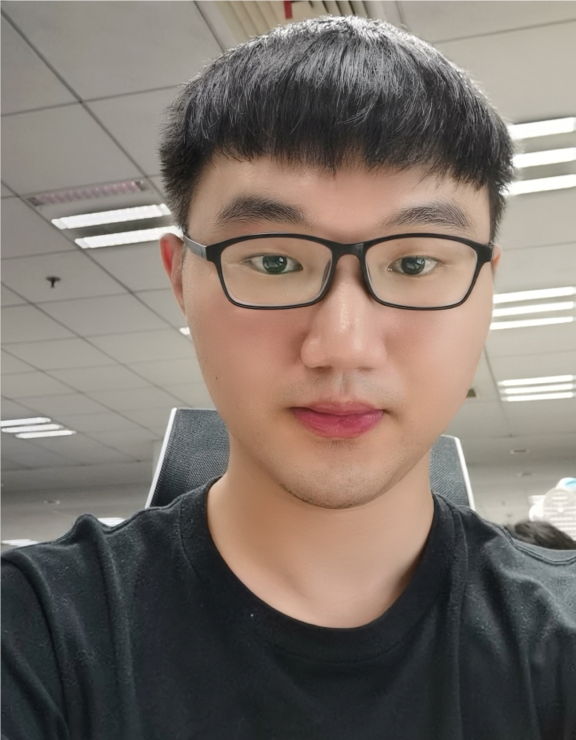}}]{Yuchen
Fang} received the B.S degree in the Department of Computer Science and
Technology from Beijing Forestry University, Beijing, China. He is currently
pursuing the M.S with the School of Computer Science (National Pilot Software
Engineering School), Beijing University of Posts and Telecommunications,
Beijing, China. His current main interests include traffic Forecasting based
on spatial-temporal data and graph neural network.
\end{IEEEbiography}
\end{document}